\PassOptionsToPackage{table}{xcolor}
\documentclass{article}

\usepackage{microtype}
\usepackage{graphicx}
\usepackage{subfigure}
\usepackage{booktabs} 

\usepackage{hyperref}

\usepackage[preprint]{icml2024}

\usepackage{amsmath}
\usepackage{amssymb}
\usepackage{mathtools}
\usepackage{amsthm}
\usepackage{multirow}
\usepackage[table]{xcolor}

\usepackage[capitalize,noabbrev]{cleveref}

\colorlet{shadecolor}{gray!20}

\theoremstyle{plain}

\theoremstyle{definition}

\theoremstyle{remark}

\usepackage[textsize=tiny]{todonotes}

\icmltitlerunning{Does Combining Parameter-efficient Modules Improve Few-shot Transfer Accuracy?}

\begin{document}

\twocolumn[
\icmltitle{Does Combining Parameter-efficient Modules Improve\\Few-shot Transfer Accuracy?}

\icmlsetsymbol{equal}{*}

\begin{icmlauthorlist}
\icmlauthor{Nader Asadi}{equal,hua}
\icmlauthor{Mahdi Beitollahi}{equal,hua}
\icmlauthor{Yasser Khalil}{hua}
\icmlauthor{Yinchuan Li}{hua}
\icmlauthor{Guojun Zhang}{hua}
\icmlauthor{Xi Chen}{hua}
\end{icmlauthorlist}

\icmlaffiliation{hua}{Noah's Ark Lab, Montreal, Canada}

\icmlcorrespondingauthor{Nader Asadi}{nader.asadi@huawei.com}

\icmlkeywords{Machine Learning, ICML}

\vskip 0.3in
]



\printAffiliationsAndNotice{\icmlEqualContribution} 


\begin{abstract}
Parameter-efficient fine-tuning stands as the standard for efficiently fine-tuning large language and vision models on downstream tasks. Specifically, the efficiency of low-rank adaptation has facilitated the creation and sharing of hundreds of custom LoRA modules, each trained on distinct data from various downstream tasks. In this paper, we explore the composability of LoRA modules, examining if combining these pre-trained modules enhances generalization to unseen downstream tasks. Our investigation involves evaluating two approaches: (a) \emph{uniform composition}, involving averaging upstream LoRA modules with equal weights, and (b) \emph{learned composition}, where we learn the weights for each upstream module and perform weighted averaging. Our experimental results on both vision and language models reveal that in few-shot settings, where only a limited number of samples are available for the downstream task, both uniform and learned composition methods result in better transfer accuracy; outperforming full fine-tuning and training a LoRA from scratch. Moreover, in full-shot settings, learned composition performs comparably to regular LoRA training with significantly fewer number of trainable parameters. Our research unveils the potential of uniform composition for enhancing transferability in low-shot settings, without introducing additional learnable parameters. 

\end{abstract}

\begin{figure}[t!]
    \centering
    \includegraphics[width=\columnwidth]{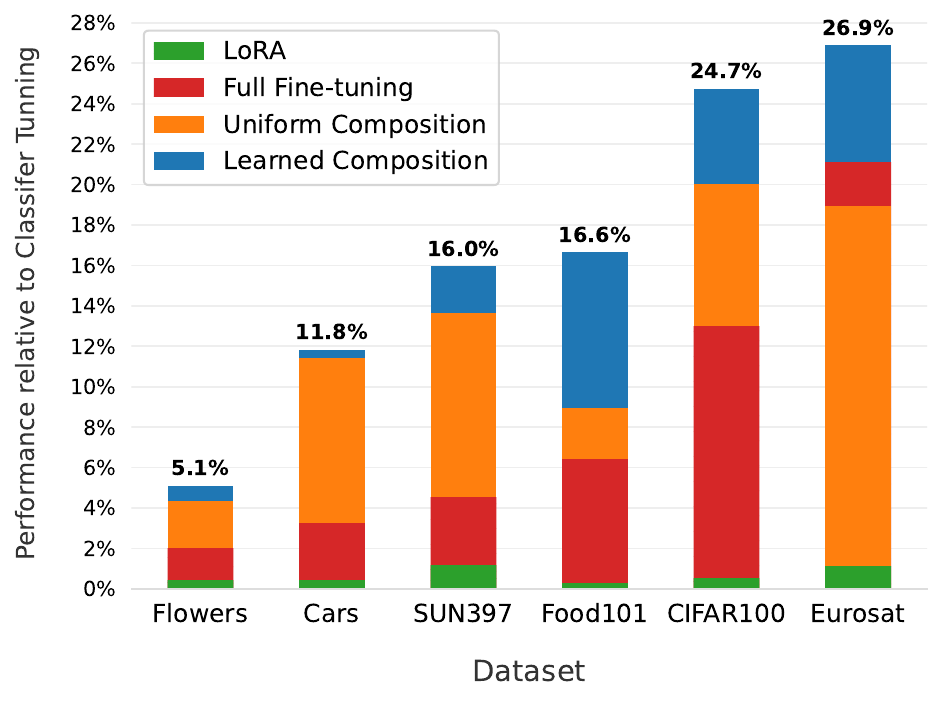}
    \vspace{-2 em}
    \caption{Performance of fine-tuning strategies relative to classifier tuning in the \emph{one-shot} transfer learning setting. Both learned~(blue) and uniform~(orange) composition methods mostly outperform regular LoRA~(green) and full fine-tuning~(red) baselines, suggesting that the linear interpolation of pre-trained LoRA modules helps few-shot transfer to an unseen downstream task. For each dataset, we use each of the rest of the dataset as an upstream task. Refer to \cref{sec:task_shift} for experiment details.}

    \label{fig:intro}
\end{figure}

\section{Introduction}\label{sec:intro}
In recent years, foundation models have demonstrated their effectiveness across a diverse set of tasks in natural language understanding, computer vision, and other fields \cite{bommasani2021opportunities}. The widespread adoption of these models, coupled with their zero-shot capability, has spurred a trend toward standardization in training models for new tasks. Both the training methodology, often involving transfer learning from popular foundational models, and the model architecture itself have conformed to established norms, typically following a few influential foundation models~\cite{dosovitskiy2020image, chung2022scaling, radford2021learning, touvron2023llama}. This standardization has given rise to numerous publicly available fine-tuned models, all sharing the same architecture. With the availability of numerous fine-tuned models derived from the same foundation model, recent studies have focused on merging multiple fine-tuned models originating from a set of upstream tasks~\cite{matena2111merging, choshen2022fusing, rame2023model, davari2023model}.

Simultaneously, due to the substantial computational cost of fine-tuning foundation models, there has been a surge in proposals for efficient \emph{adapter} modules, enabling \emph{parameter-efficient fine-tuning} of these models~\cite{lester2021power, hu2021lora, liu2022few}. Notably, low-rank adaptation (LoRA)~\cite{hu2021lora} has emerged as an efficient fine-tuning technique. LoRA involves adding and training lightweight modules to a frozen pre-trained model, achieving good performance on the downstream task. By alleviating high memory demands and computational costs, LoRA has become the standard for fine-tuning Large Language Models (LLMs), diffusion models, and vision transformers across various downstream tasks \cite{dettmers2023qlora, xu2023baize, gandikota2023concept, shah2023ziplora}. LoRA's efficiency has empowered developers to create and share custom models trained on their unique data for new tasks, resulting in the availability of hundreds of publicly accessible LoRA modules tailored for diverse downstream tasks.

This paper explores the possibility of leveraging pre-trained LoRA modules for efficient fine-tuning on a new task. Inspired by the literature on model merging~\cite{wortsman2022model, choshen2022fusing, ilharco2022editing}, we explore the composability of LoRA modules, examining whether knowledge from multiple upstream tasks can be combined for tackling new tasks. Specifically, we aim to answer this question: Does combining pre-trained LoRA modules enhance transfer accuracy on unseen tasks?

To answer this question, we adopt a few-shot transfer setting, where we train LoRA modules on diverse upstream tasks and subsequently evaluate various composition strategies on a downstream task with a limited number of samples. We evaluate two combining strategies: (a) \emph{uniform composition}, where upstream LoRA modules are averaged with equal weights, and (b) \emph{learned composition}, where we learn weights for each upstream module and perform weighted averaging.
In real-world scenarios, the upstream and downstream tasks can be entirely disentangled, originating from distinct datasets, diverse domains, or even different parts of the same dataset. To comprehensively assess the impact of this disentanglement on the performance of composition methods, we examine three scenarios wherein upstream tasks are dissociated from the downstream tasks: (a) \emph{label shift}, where upstream tasks and the downstream task each possess distinct labels from the same dataset; (b) \emph{domain shift}, involving the downstream task accessing a unique domain from the same dataset; and (c) \emph{task shift}, wherein both upstream and downstream tasks have exclusive datasets featuring different tasks.

Our findings in vision and language models demonstrate that the combination of pre-trained LoRA modules enhances generalization in a few-shot setting. Specifically, both uniform and learned composition methods yield superior transfer accuracy, outperforming full fine-tuning and training a LoRA from scratch as shown in Figure \ref{fig:intro}. Furthermore, our results indicate that as the number of samples in the downstream task increases, learned composition maintains performance on par with full fine-tuning and regular LoRA training while utilizing significantly fewer trainable parameters. Lastly, a comparison of the learned and uniform composition reveals that as the number of samples in the downstream task increases, the performance gap between learned and uniform composition widens, with learned composition consistently delivering superior results.


\begin{figure*}[t!]
    \centering
    \includegraphics[width=0.99\textwidth]{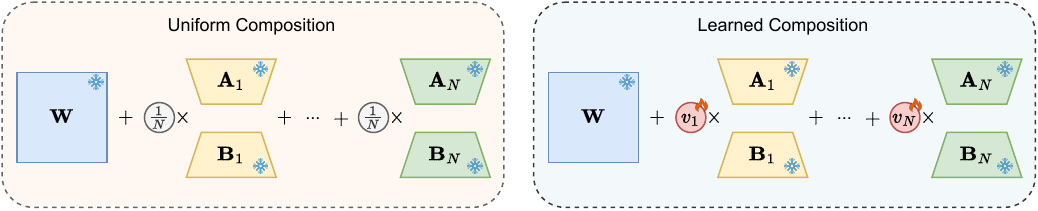}
    \vspace{-0.5 em}
    \caption{\textbf{Method overview.} We start with a foundational model that has undergone LoRA fine-tuning on various tasks. During the few-shot adaptation phase, for each layer of the model, we apply a \emph{uniform} (left) or \emph{learned} (right) weighted averaging over the pre-trained upstream LoRA weights.}
    \vspace{-0.5 em}
    \label{fig:merging_methods}
\end{figure*}

\section{Related Works}\label{sec:background}

\paragraph{Model Merging}
These studies aim to bring together separately trained models to either mimic the advantages of multi-task learning or enhance the merged model's performance on new data distributions.
\citealt{wortsman2022model} showed that models fine-tuned with different hyperparameters on the same dataset can be merged through a weighted average, resulting in a superior-performing model. However, they didn't explore merging models from different tasks.
\citealt{matena2111merging} focused on merging multiple fine-tuned models derived from the same pre-trained model but trained on diverse datasets. The merging process involves a weighted average determined by the Fisher information matrix. However, computing the Fisher information matrix becomes computationally expensive with an increasing number of models.
\citealt{choshen2022fusing} demonstrated that fusing multiple fine-tuned models surpasses the original pre-trained model in adapting to a new target domain. \citealt{rame2023model} took a step further, using multiple fine-tunings of the same foundation model as initializations for parallel fine-tunings on target tasks, showing that the averaged model achieves the best performance in out-of-distribution (OOD) generalization.
\citealt{ilharco2022editing} introduced \textit{Task Vectors}, calculated by subtracting fine-tuned model weights from their pre-trained initialization. They showed that scaled averaging of fine-tuned models starting from a pre-trained model creates a single multi-task model. Following the Task Vectors idea, \citealt{davari2023model} proposed a masked averaging method to combine several fine-tuned models into a single multi-task model.

\paragraph{Adapters} 
To address the computational and memory challenges associated with fine-tuning foundation models, there has been a recent emphasis on developing parameter-efficient modules known as \emph{adapters} to efficiently fine-tune pre-trained models on downstream tasks.
Early works \cite{rebuffi2017learning, bapna2019simple, lin2020exploring} proposed to use small trainable feed-forward modules inserted into the frozen pre-trained model. Other works \cite{guo2020parameter, sung2021training} suggested selectively fine-tuning only a subset of parameters. Some other works \cite{li2021prefix, lester2021power} proposed to add some learned embeddings to the model’s input or activations to induce it to perform a task. \citealt{hu2021lora} tries to make fine-tuning more efficient by representing the weight updates with two smaller matrices, \textit{i.e.} a low-rank weight matrix with similar input and output dimensionality as the original weight.

\paragraph{Adapter Composition}
Due to the efficiency of parameter-efficient fine-tuning methods, the adoption of adapter-based techniques for fine-tuning large pre-trained models has become prevalent. Therefore, recent efforts have focused on combining multiple adapters for multi-task learning.
In a study by \citealt{ponti2023combining}, the multi-task challenge is tackled by softly sharing a set of adapter parameters across tasks. With $N$ representing the number of adapters and $T$ denoting the multi-task datasets, they devise a model where $N \ll T$ by learning a set of adapters and a routing function.
As a follow up work, \citealt{caccia2023multi}, suggests employing a multi-head attention operation as the routing function introduced by \cite{ponti2023combining}, aiming to enhance model performance with a slight increase in parameter count.
As opposed to these works, we aim to study the benefits of merging pre-trained adapters for few-shot transfer to unseen downstream tasks.
\citealt{huang2023lorahub} proposes the idea of utilizing available LoRA modules to achieve adaptable performance on unseen tasks. They use a black-box gradient-free approach to select a handful of adapter modules from a hub of thousands. Although the idea of utilizing available LoRAs is close to the exploratory objectives of our work, their adapter merging approach is too noisy and does not demonstrate the effect of simple merging methods on adapters.  \looseness=-1


\section{Problem Definition}\label{sec:problem_definition}
\paragraph{Low-Rank Adaption~(LoRA)}
Starting from a pre-trained model $\Theta_0$, regular fine-tuning learns a different set of parameters $\Theta$ for each downstream task with $|\Theta| = |\Theta_0|$. 
Instead, LoRA tries to learn a set of task-specific parameters $\Delta\Theta$ with  a much smaller-sized set of parameters compared to $\Theta_0$ with $|\Delta\Theta| \ll |\Theta_0|$.
Given a pre-trained weight matrix $\mathbf{W}_0 \in \mathbb{R}^{d \times c}$ of the pre-trained model $\Theta_0$, LoRA adds a trainable low-rank decomposition matrix $\Delta\mathbf{W}$ as adapter modules to the original weight matrix $\mathbf{W}_0$: 
\begin{equation*}
    \hat{\mathbf{W}} = \mathbf{W}_0 + \alpha \Delta\mathbf{W}, 
\end{equation*}
where $\Delta\mathbf{W} = \mathbf{A}\mathbf{B}^{\top}$ represents a low-rank matrix with rank $r \ll min(d,k)$ where $\mathbf{A} \in \mathbb{R}^{d \times r}$, $\mathbf{B}^{\top} \in \mathbb{R}^{r \times c}$ and $\alpha$ is a weighting coefficient. 
Then finding the value of $\Delta\Theta$ can be formulated as the standard maximum-likelihood training with cross-entropy for conditional language modeling:
\begin{equation*}
    \max _{\Delta\Theta} \sum_{(x, y) \in \mathcal{D}} \sum_{t=1}^{|y|} \log \left(P_{\Theta_0+\Delta \Theta}\left(y_t \mid x, y_{<t}\right)\right)
\end{equation*}

\paragraph{Few-shot Transfer Setup}
The goal is to build a single model, personalized for a novel domain utilizing the pre-trained LoRA modules from upstream domains. To evaluate the usefulness of the upstream pre-trained LoRA modules for the downstream tasks, we consider the few-shot transfer learning setting.  Assuming that we have $N$ distinct upstream tasks denoted as $\mathbb{T} = \{\mathcal{T}_1, ..., \mathcal{T}_N\}$ each having a set of trained LoRA modules. We evaluate the performance of several merging approaches of these upstream modules on a new unseen target domain $\mathcal{T}^{\prime} \notin \mathbb{T}$. 
Each upstream domain $\mathcal{T}_n$ is defined by a set of data points $\mathcal{X}_n$, a set of ground truth labels $\mathcal{Y}_n$, and a distribution $\mathcal{D}_n$ over $\mathcal{X}_n$ and $\mathcal{Y}_n$. Similarly, the target domain $\mathcal{T}^{\prime}$ is defined by a set of data points $\mathcal{X}^{\prime}$, a set of ground truth labels $\mathcal{Y}^{\prime}$, and a distribution $\mathcal{D}^{\prime}$ over $\mathcal{X}^{\prime}$ and $\mathcal{Y}^{\prime}$.
The few-shot learning task in the target domain consists of a very small subset of training data or \emph{support} set from $\mathcal{D}^{\prime}$:
\begin{equation*}
    \mathcal{S}^{K} = \{(x_k, y_k)\}^{K}_{k=1} \sim \mathcal{D}^{\prime} , \quad y_i \in \mathcal{Y}^{\prime}
\end{equation*}
where $K$ represents the number of adaptation samples per class, used to fine-tune the model on downstream task $\mathcal{T}^{\prime}$. For the evaluation, we use all of the samples in the test or \emph{query} set of the downstream dataset $\mathcal{D}^{\prime}$.

\paragraph{Objective}
Assume for each upstream task $\mathcal{T}_n \in \{\mathcal{T}_1, ..., \mathcal{T}_N\}$, we have a set of fine-tuned LoRA modules denoted as $\Delta\mathbf{W}_n \in \{\Delta\mathbf{W}_1, ..., \Delta\mathbf{W}_N\}$. The objective is to find a combination of the upstream LoRA modules using the $K$ samples in support set $\mathcal{S}^{K}$ of the unseen task $\mathcal{T}^{\prime}$ to improve the performance on the query or test set. 
For the language modeling experiments, we follow the procedure from \cite{raffel2020exploring} and formulate each task as a text-to-text problem, enabling standard maximum-likelihood training with a cross-entropy loss.  


\begin{figure}[t!]
    \centering
    \includegraphics[width=0.99\columnwidth]{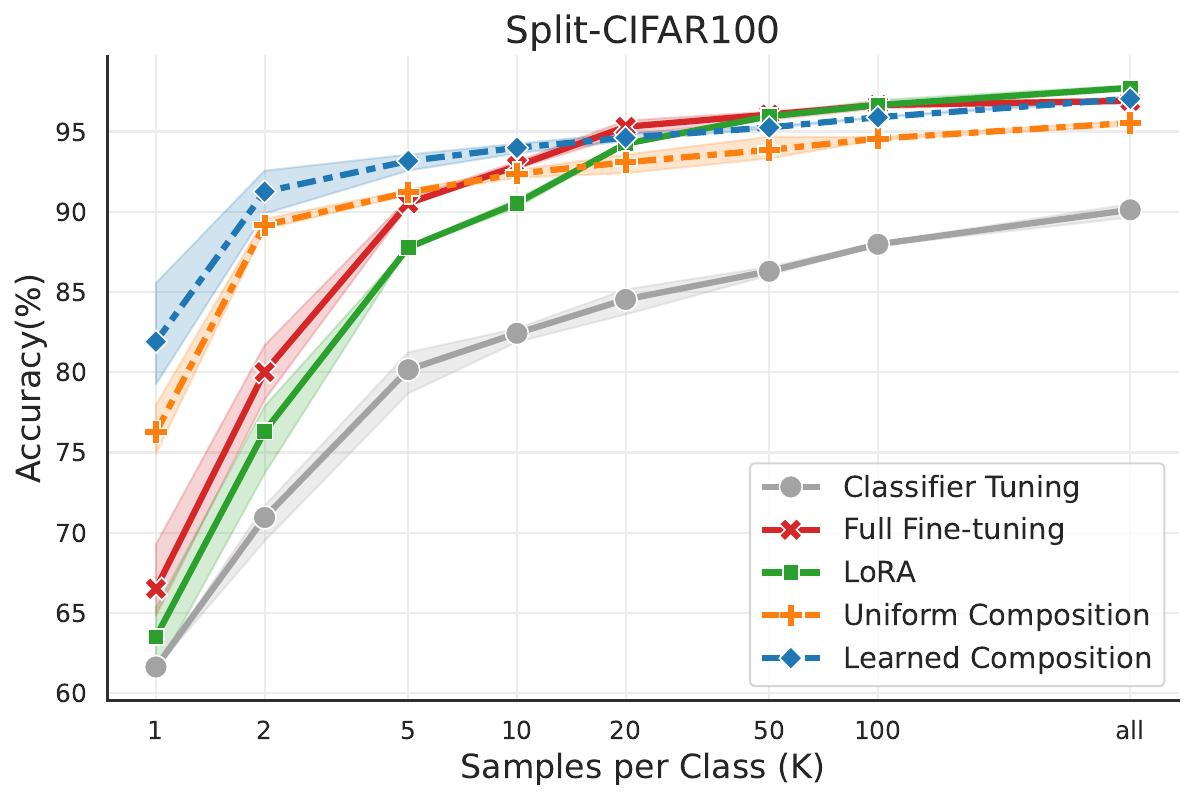}
    \vspace{-1 em}
    \caption{\textbf{Label shift results.} Few-shot transfer results with different numbers of adaptation samples. We can observe that \emph{uniform} and \emph{learned} composition methods consistently beat other baselines with fewer number of adaptation samples.}
    \vspace{-1 em}
    \label{fig:cifar100}
\end{figure}

\section{Combination Methods}
This section highlights different strategies for combining pre-trained LoRA modules. Our objective is to effectively merge these pre-trained low-rank modules, which were originally trained on disjoint auxiliary tasks, to enhance performance in a new unseen downstream task with a limited number of samples. We consider two major recipes for merging the pre-trained adapters as shown in \cref{fig:merging_methods}: \textit{uniform} and \textit{learned}.   \looseness=-1

\subsection{Uniform Composition} 
We begin with a pre-trained foundation model $\Theta_0$ that has undergone fine-tuning with LoRA for various auxiliary tasks. Denoting each weight matrix of the foundation model as $\mathbf{W}_0$, LoRA fine-tuning adds a low-rank matrix $\Delta \mathbf{W}_n$ for the auxiliary task $\mathcal{T}_n$.
The uniform composition is constructed by averaging all fine-tuned LoRA modules as:
\begin{equation*}
    \setlength{\abovedisplayskip}{2pt}
    \setlength{\belowdisplayskip}{2pt}
    \hat{\mathbf{W}} = \mathbf{W}_0 + \frac{1}{N} \sum_{n=1}^{N} \Delta \mathbf{W}_n.
\end{equation*}

\subsection{Learned Composition}
We also explore a more advanced learned composition recipe that optimizes LoRA interpolation weights by gradient-based minibatch optimization. The learned composition allows us to determine a specific interpolation of LoRA modules that best suits the downstream task $\mathcal{T}^{\prime}$.
It's worth noting that this procedure requires loading all LoRA weights into memory simultaneously. However, due to the low dimensionality of LoRA parameters, this operation is feasible, unlike learning interpolation parameters across large fine-tuned models~\cite{wortsman2022model}.
More specifically, we learn a weighting vector parameter $\mathbf{v} \in \mathbb{R}^{N}$, where $v_n \in \mathbb{R}$ denotes to the $n$-th element of $\mathbf{v}$ representing the weighting coefficient for the adapter of upstream task $n$ as follows:
\begin{equation*}
    \setlength{\abovedisplayskip}{2pt}
    \setlength{\belowdisplayskip}{2pt}
    \hat{\mathbf{W}} = \mathbf{W}_0 + \sum_{n=1}^{N} \hat{v}_n \Delta \mathbf{W}_n,
\end{equation*}
where $\hat{v}_n = \frac{e^{v_n}}{\sum_{j=1}^{N} e^{v_j}}$ is the softmax operation applied on the weighting vector $\mathbf{v}$. 
Note that both the uniform and learned composition methods are independent of the size of the upstream LoRA rank.


\begin{figure*}[t!]
    \begin{minipage}{0.49\textwidth}
        \includegraphics[width=1.0\textwidth]{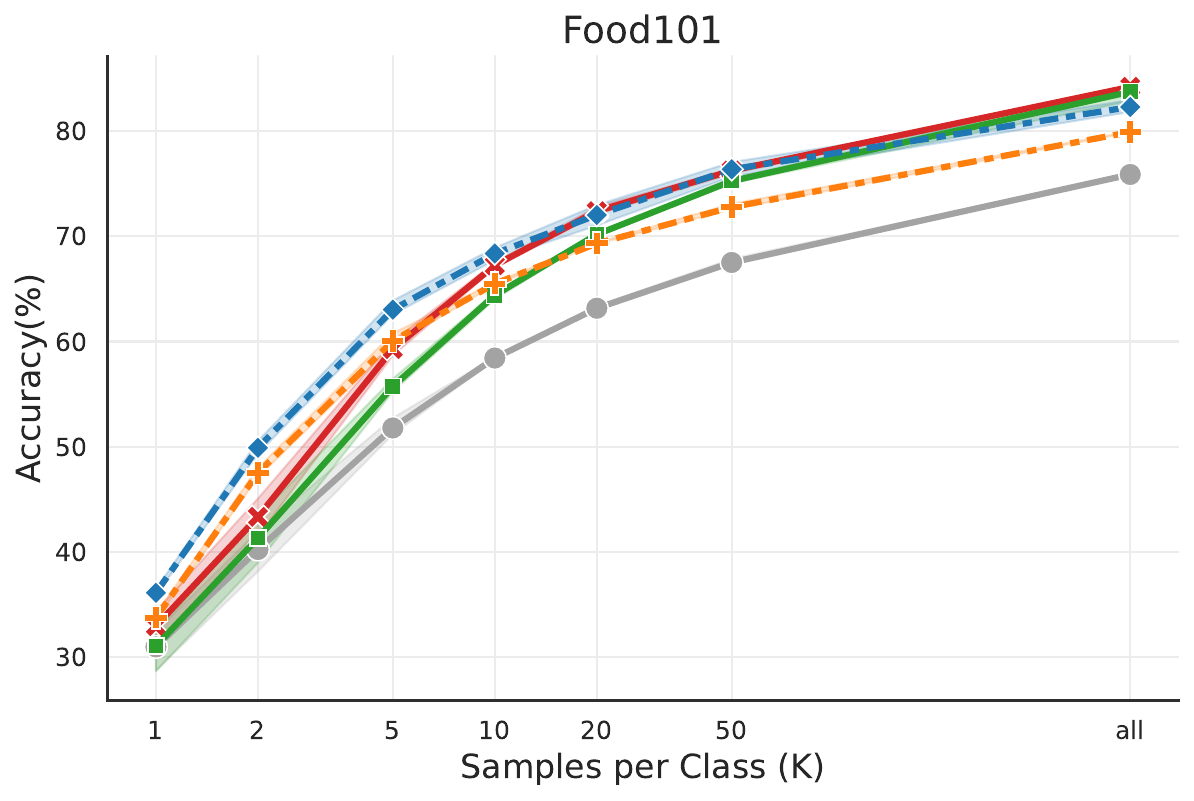}
    \end{minipage}\hfill
    \begin{minipage}{0.49\textwidth}
        \includegraphics[width=1.0\textwidth]{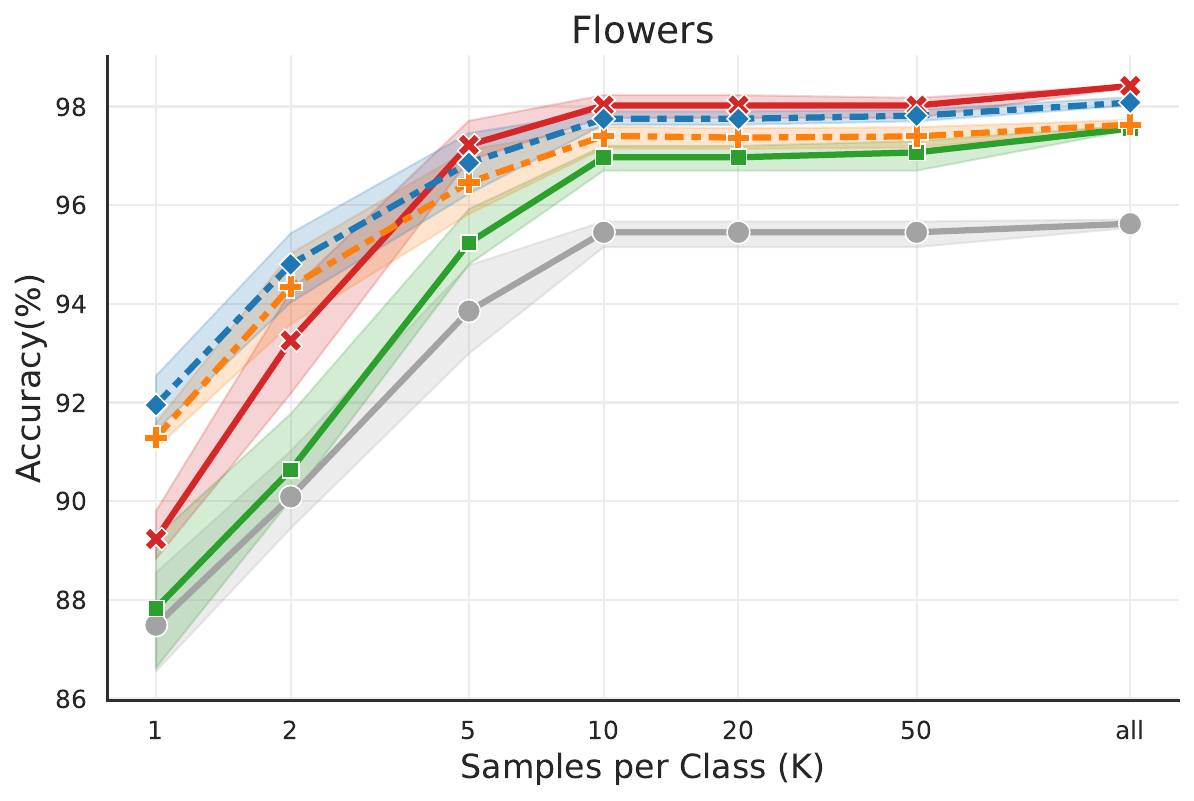}
    \end{minipage}\hfill
    \begin{minipage}{0.49\textwidth}
        \includegraphics[width=\textwidth]{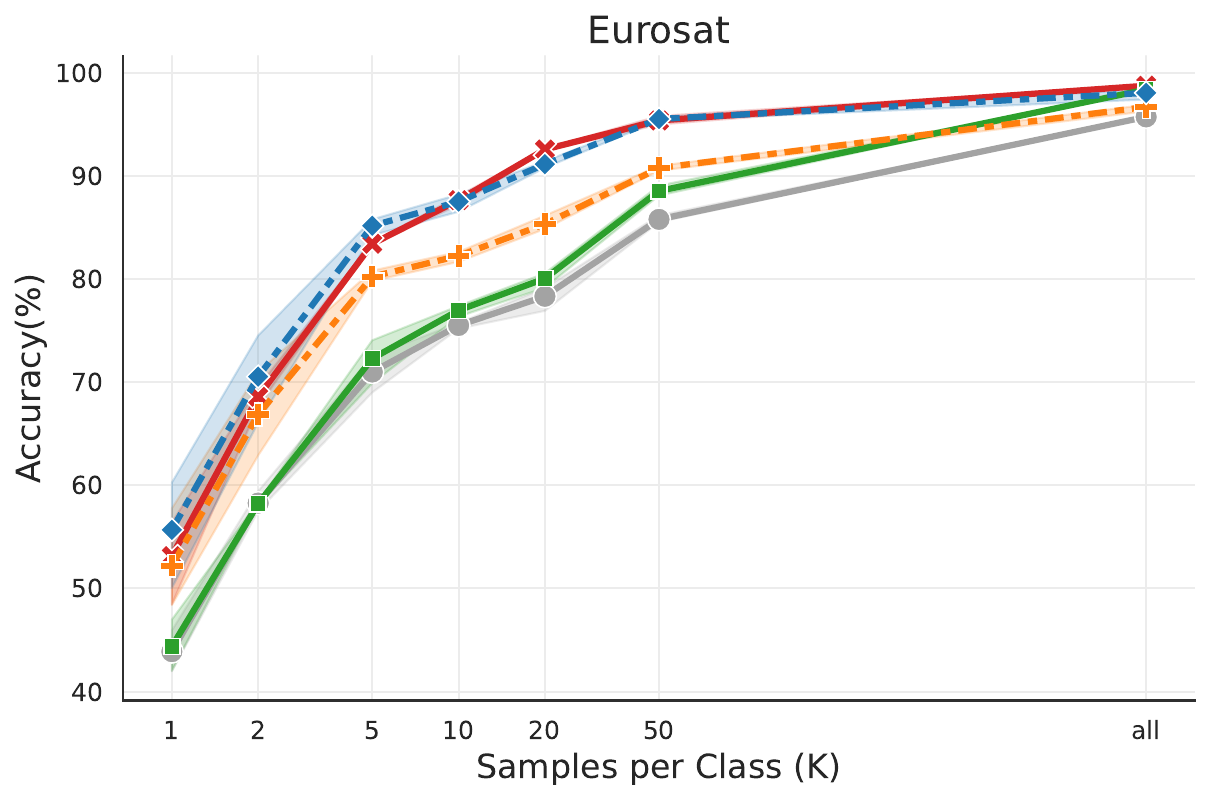}
    \end{minipage}\hfill
    \begin{minipage}{0.49\textwidth}
        \includegraphics[width=1.0\textwidth]{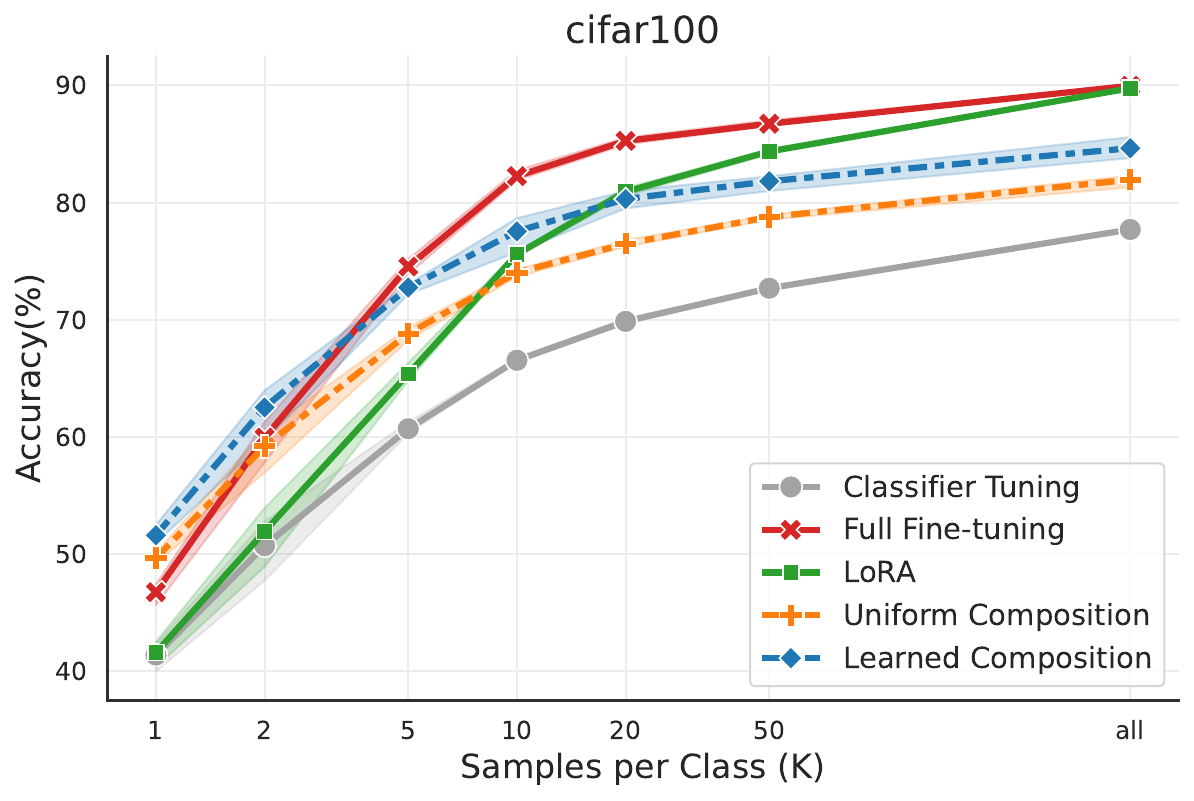}
    \end{minipage}\hfill
    \caption{\textbf{Task shift results~(vision).} Few-shot transfer results with different number of adaptation samples. We observe that both uniform and learned composition methods significantly improve the performance with few number of adaptation samples, while maintaining comparable performance against regular LoRA fine-tuning in the full-shot scenario.}
    \label{fig:task_shift}
\end{figure*}

\section{Experiments}\label{sec:experiments}

Our experimental analysis aims to answer the following question: Does combining pre-trained LoRA modules enhance transfer accuracy on new tasks?
We first explain our benchmark setup, then try to answer this question based on our observations.

\begin{table}[h!]
    \small
    \centering
     \begin{tabular}{lllll}
    \toprule
    Dataset         & \# Train          & Task & Setting  \\
    \midrule
    Spli-CIFAR100    & 50,000 & Vision & Label Shift \\  
    \midrule
    PACS  & 1,336 &Vision  & Covariate Shift \\  
    \midrule
    CIFAR100  & 50,000 & Vision & Task Shift \\
    Food101 & 75750 & Vision & Task Shift \\
    Stanford Cars  & 12948 & Vision & Task Shift \\
    SUN397     & 108,754&Vision  & Task Shift \\
    Flowers   & 7,169 &Vision & Task Shift  \\
    Eurosat   & 27,000 &Vision  & Task Shift \\
    \midrule    
     SciQ & 13,679 & NLP &Task Shift \\
     CommonSense & 12,102  & NLP &Task Shift   \\
     QuAIL & 15,000  & NLP &Task Shift   \\
     ARC & 7,787  & NLP &Task Shift    \\
    \bottomrule
  \end{tabular}
  \caption{Summary of datasets used for each evaluation setting.\vspace{-1em}}
  \label{tbl:datasests}
\end{table}

\paragraph{Setup}
For all of the experiments, learning takes place in three phases. The first phase is considered as pre-training of the foundation model $\Theta_0$. For all of the vision experiments, we considered ViT-base~\cite{dosovitskiy2020image} with a patch-size of 32$\times$32, pre-trained on ImageNet-21K~\cite{ridnik2021imagenet}. For the NLP experiments, we use the pre-trained Flan-T5 large~\cite{chung2022scaling}; refer to the original paper for more information.
The second phase consists of fine-tuning a set of LoRA adapters, on the set of disjoint auxiliary tasks. We refer to this phase as the upstream training stage. The third and final phase consists of a few-shot adaptation to a new unseen task.
For the vision task, we focus on the image classification problem, reporting \emph{top-1 accuracy} as our evaluation metric. For the NLP task, we focus on multi-choice question answering problems, reporting \emph{exact match} as the evaluation metric. We summarize the datasets in \cref{tbl:datasests}.

\paragraph{Hyperparameter selection}
In all our experiments, we train LoRA with a rank of 16 for the query, key, and value weight matrices. This practice applies to both the ViT-base and Flan-T5 models unless stated otherwise. In our vision experiments using the ViT-base, we warm up the classifier head for 50 epochs at the beginning of training. For each method, optimal hyperparameters were selected via a grid search performed on the validation set. The selection process was done on a per-dataset basis, where we picked the configuration that maximized the accuracy averaged over different settings. We report performance with the mean and standard deviation, calculated over three random seeds.  


\begin{table*}[t!]
  \centering
  \begin{tabular}{l lllllc}
    \toprule
    \multirow{2}{*}{Method} &   \multicolumn{5}{c}{ARC-Challenge} & \\
    \cmidrule{2-7}
    & \multicolumn{1}{c}{K=1} & \multicolumn{1}{c}{K=2} & \multicolumn{1}{c}{K=3} & \multicolumn{1}{c}{K=4} & \multicolumn{1}{c}{K=5}  & \multicolumn{1}{c}{$|\Theta|$}\\
    \midrule

    Zero-shot & 49.31\tiny{ $\pm$0.0} & 49.31\tiny{ $\pm$0.0} & 49.31\tiny{ $\pm$0.0} & 49.31\tiny{ $\pm$0.0} & 49.31\tiny{ $\pm$0.0} & 0\\
    Full Fine-tuning & 57.05\tiny{ $\pm$}1.02 & 59.06\tiny{ $\pm$}0.98 & \underline{59.71}\tiny{ $\pm$0.42} & \textbf{60.94}\tiny{ $\pm$}0.58 & \textbf{62.13}\tiny{ $\pm$}0.61 & 787M\\
    LoRA & 56.48\tiny{ $\pm$1.20} & 58.44\tiny{ $\pm$0.93} & 58.95\tiny{ $\pm$0.34} & 59.47\tiny{ $\pm$0.42} & 60.16\tiny{ $\pm$0.65} &  471M\\
    \midrule
    Uniform Composition & \underline{59.11}\tiny{ $\pm$0.0} & \underline{59.11}\tiny{ $\pm$0.0} & 59.11\tiny{ $\pm$0.0} & 59.11\tiny{ $\pm$0.0} & 59.11\tiny{ $\pm$0.0} & 0\\
    Learned Composition & \textbf{59.40}\tiny{ $\pm$0.87} & \textbf{59.97}\tiny{ $\pm$0.61} & \textbf{60.39}\tiny{ $\pm$0.30} & \underline{60.67}\tiny{ $\pm$0.39} & \underline{61.12}\tiny{ $\pm$0.43} & 1.6K\\

   \bottomrule
  \end{tabular}
    \caption{\textbf{Task shift results~(NLP).} We can observe that the uniform composition method improves the transfer performance of Flan-T5 large by 9.8\% in the zero-shot setting. Also, the learned composition method beats the full and LoRA fine-tuning baselines with less than 3 and 5 adaptation samples respectively. Note that $|\Theta|$ represents the number of trainable parameters.}
  \label{tab:arc}
\end{table*}

\begin{table*}[t!]
  \centering
  \begin{tabular}{l llllll}
    \toprule
    \multirow{2}{*}{Method} &   \multicolumn{6}{c}{PACS (Art painting)}\\
    \cmidrule{2-7}
    & \multicolumn{1}{c}{K=1} & \multicolumn{1}{c}{K=5} & \multicolumn{1}{c}{K=10} & \multicolumn{1}{c}{K=50} & \multicolumn{1}{c}{K=100} & \multicolumn{1}{c}{K=all} \\
    \midrule

    Classifier Tuning   &46.37\tiny{ $\pm$ 12.00} &69.35\tiny{ $\pm$ 2.50}  &73.02\tiny{ $\pm$ 2.19}  &82.88\tiny{ $\pm$ 0.97}  &84.68\tiny{ $\pm$ 0.37}   & 86.30\tiny{ $\pm$ 0.24}\\
    Full Fine-tuning &50.28\tiny{ $\pm$ 9.88}  &\underline{79.05}\tiny{ $\pm$ 2.83}  &\textbf{83.45\tiny{ $\pm$ 1.99}}  &\textbf{92.58\tiny{ $\pm$ 0.14}}  &\textbf{93.64\tiny{ $\pm$ 0.42}}  &\textbf{95.43\tiny{ $\pm$ 0.98}}\\
    LoRA &44.66\tiny{ $\pm$ 12.00}  &77.51\tiny{ $\pm$ 2.48} &\underline{83.29}\tiny{ $\pm$ 1.15}  &\underline{91.93\tiny{ $\pm$ 0.64}} &\underline{93.07\tiny{ $\pm$ 0.85}}  & \underline{93.88 \tiny{ $\pm$ 0.24}}\\

    \midrule
    Uniform Composition &\underline{56.88}\tiny{$\pm$ 11.02}  &78.73\tiny{$\pm$ 2.30}  &81.09\tiny{$\pm$ 1.92}  &86.47\tiny{$\pm$ 0.42}  &88.02\tiny{$\pm$ 1.21}  &90.06\tiny{$\pm$ 0.30}    \\
    Learned Composition &\textbf{57.21\tiny{$\pm$ 11.56}}  &\textbf{79.54\tiny{$\pm$ 2.31}}  &81.25\tiny{$\pm$ 1.75}  &86.55\tiny{$\pm$ 0.80}  &88.43\tiny{$\pm$ 0.99}  &90.87\tiny{$\pm$ 0.42}    \\

   \bottomrule
  \end{tabular}
  \caption{\textbf{Covariate shift results}. Utilizing the upstream modules from Photo~(P), Cartoon~(C), and Sketch~(S) domains, uniform and learned composition methods beat the performance of other baselines with less than 5 adaptation samples.}
  \label{tab:pacs_art}
\end{table*}

\subsection{Label Shift}\label{sec:label_shift}
We begin our study with label shift as the first scenario to explore the benefits of uniform and learned LoRA composition under the few-show transfer setting. In this experiment, we consider Split-CIFAR100~\cite{krizhevsky2009learning}, comprised of 4 subsets, each containing a disjoint set of 25 labels. We consider 3 of these as the upstream~(seen) tasks, and the 4$^{\text{th}}$ as the downstream~(unseen) task. \cref{fig:cifar100} demonstrates the performance of different methods under the few shot transfer setting. The x-axis represents the number of support samples per class for adapting to the downstream task. We can observe that specifically in the low data regime, with less than 10 samples per class, the learned composition is able to find a combination of upstream expert adapters for the downstream task. Interestingly, even in the full-shot setting, the learned composition method manages to perform comparably to the conventional LoRA fine-tuning approach, all while maintaining an order of magnitude fewer trainable parameters.

\subsection{Task Shift}\label{sec:task_shift}
To further explore the benefits of merging pre-trained LoRA modules for cross-task generalization, we conducted experiments under task shift for both vision and natural language understanding domains.

\textbf{Vision Results   }
For our vision experiments, we consider using a subset of the 6 datasets, as used in previous work \cite{kornblith2019better}. We assessed few-shot transfer accuracy across Stanford Cars~\cite{krause20133d}, Food101~\cite{bossard2014food}, Sun397~\cite{xiao2010sun}, Eurosat~\cite{helber2019eurosat}, Flowers~\cite{nilsback2008automated}, and CIFAR100~\cite{krizhevsky2009learning}. Our task shift experiments were conducted in 6 rounds, where each round considered one of these datasets as the downstream task and the others as the upstream tasks. The results can be seen in \cref{fig:task_shift} for Food101, Flowers, EuroSat, and CIFAR100 as the downstream tasks. Note that the few-shot downstream adaptation is the third phase of our experimental setup, with the second phase involving the auxiliary training of upstream adapters on the remaining datasets mentioned earlier.

From \cref{fig:task_shift}, we can observe that the uniform composition still improves the model's performance when there's very little data available. Specifically, in all cases, uniform composition outperforms both LoRA and regular fine-tuning methods when we have only 1 or 2 samples per class. On the other hand, the learned composition not only performs better in low-data situations but also maintains good transfer performance across the entire spectrum. Interestingly, the learned composition performs nearly as well as LoRA fine-tuning in a scenario with plenty of data, while having a considerably smaller number of trainable parameters. This observation answers our initial question: pre-trained LoRA modules on different tasks can indeed enhance the model's performance on new, unseen tasks.

\textbf{NLP Results   }
For the NLP experiments, we consider a subset of CrossFit benchmark~\cite{ye2021crossfit}. We focus on multi-choice question answering problem using SciQ~\cite{bhakthavatsalam2021think}, CommonSense~\cite{welbl2017crowdsourcing}, QuAIL~\cite{rogers2020getting}, and ARC~\cite{bhakthavatsalam2021think} datasets. We evaluate the few-shot transfer accuracy of Flan-T5 large, reporting the exact match~(EM) as our evaluation metric. \cref{tab:arc} presents our results on ARC dataset. We consider SciQ, CommonSense, and QuAIL as the upstream tasks. Note that the uniform composition method is also evaluated in zero-shot setting. We can observe that the uniform merging of the upstream adapters significantly improves the zero-shot performance of the model. Additionally, the learned composition outperforms regular LoRA fine-tuning when less than five samples are available for training.

\subsection{Covariate Shift}

We also evaluate the transferability of pre-trained LoRA modules in covariate shift settings, where upstream tasks and downstream tasks each have unique domains. To evaluate this setting, we utilize PACS dataset~\cite{li2017deeper}, where we consider the three upstream domains as Photo (P), Cartoon (C), and Sketch (S); and the downstream domain as Art Painting (A) images. The results can be found in \cref{tab:pacs_art}. We can see that our observation is consistent with previous experiments. Uniform and learned composition methods consistently beat all of the other baselines by a considerable margin. In \cref{sup:covariate_shift}, we also report the results on the Sketch (S) domain as the downstream and Photo (P), Cartoon (C), and Art Painting (A) as upstream domains.  \looseness=-1

\begin{figure}[t!]
    \centering
    \includegraphics[width=0.99\columnwidth]{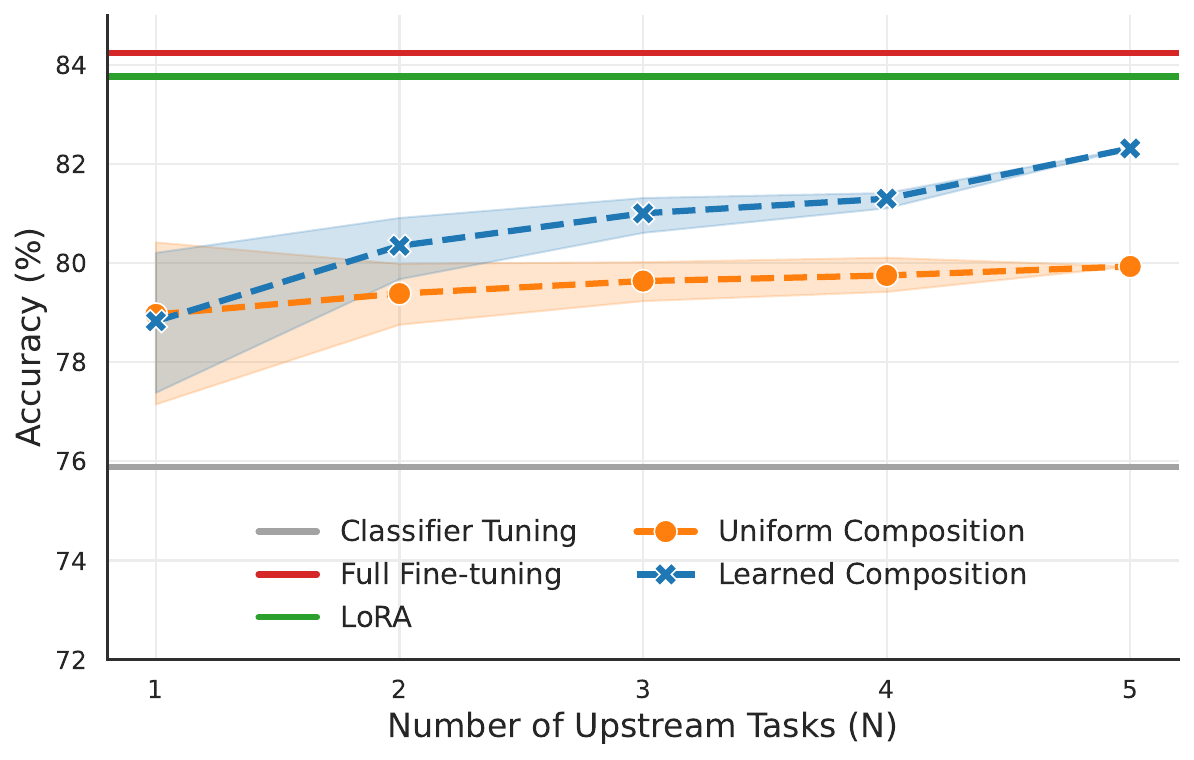}
    \caption{\textbf{Effect of scaling number of upstream tasks.} The results on Food101 dataset in full-shot ($K$ = all) scenario indicate that, as the number of pre-trained upstream modules increases, the learned composition of these modules significantly enhances the model’s performance.}
    \vspace{-1 em}
    \label{fig:memory}
\end{figure}

\section{Analysis}\label{sec:analysis}

\subsection{Scaling the Number of Upstream Tasks ($N$)}
In this section, we explore how the number of pre-trained upstream LoRA modules affects the transfer accuracy to a new dataset under task shift setting. \cref{fig:memory} presents our results on Food101 dataset in full-shot ($K$ = all) scenario.
Our findings indicate that, as the number of pre-trained upstream modules increases, the learned composition of these modules significantly enhances the model's performance, narrowing the performance gap compared to full-finetuning even in scenarios where all data is available.
Notably, when utilizing only one upstream module, we observe a considerable variation in model performance. This variation stems from differences in the usefulness of individual upstream modules for the downstream task.
Interestingly, our analysis reveals that the performance achieved by the learned composition of all five upstream LoRA modules surpasses that of the best individual module. This suggests that leveraging a linear combination of LoRA modules can notably improve transfer accuracy across diverse tasks. In \cref{sec:cka}, we delve into how the learned composition effectively selects relevant upstream modules, shedding further light on the mechanism driving these improvements. 

\begin{figure}[t!]
\centering
    \includegraphics[width=0.99\columnwidth]{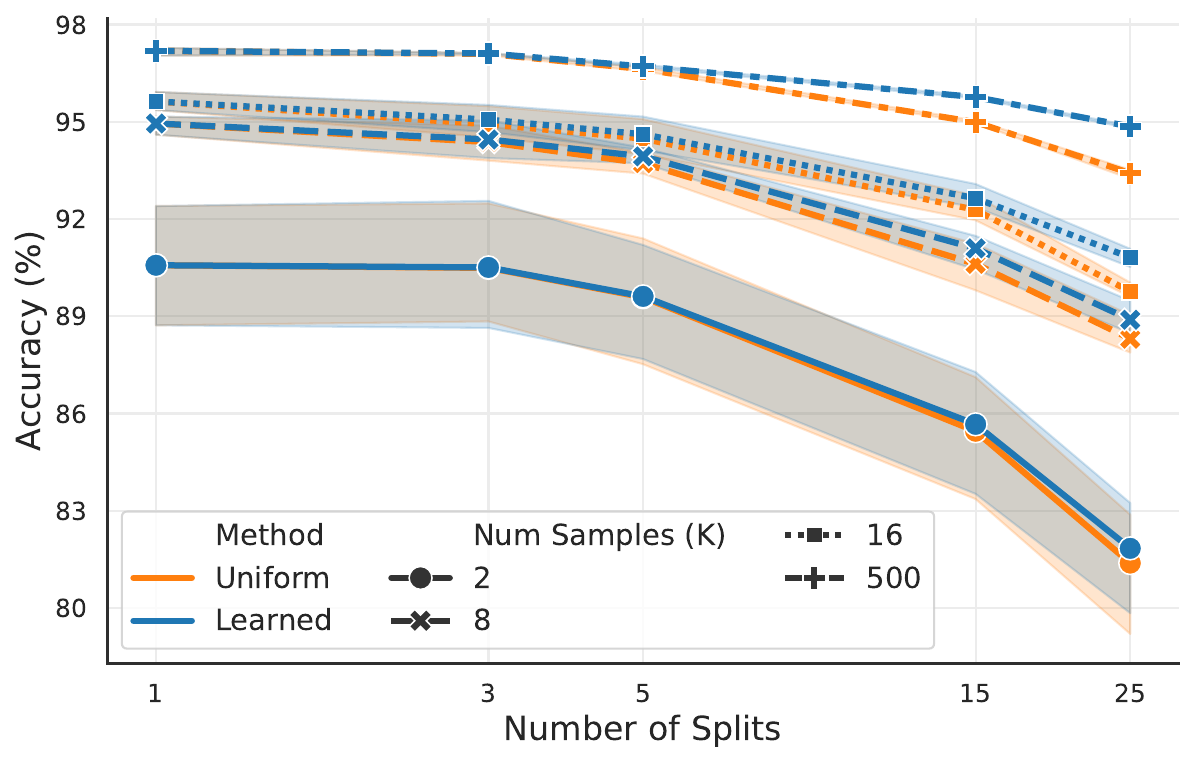}
    \caption{\textbf{Effect of split-size.} As the number of Split-CIFAR100 subsets increases in label shift setting, the performance gap between uniform and learned composition methods increases as well.}
    \label{fig:long_seq}
\end{figure}

\subsection{Effect of Split Size}
In this section, we aim to do an ablation study on the effect of split-size in label shift setting on the performance of composition methods.
More specifically, we split the Split-CIFAR100 dataset into 2 sets of 25 labels for the downstream task and 75 labels to allocate to the downstream tasks. Then, we further split those 75 labels into $N=\{1, 3, 5, 15, 25\}$ disjoint upstream tasks, and report the results in Figure \ref{fig:long_seq}. 
We can observe that as the number of upstream tasks increases, \textit{i.e.} more LoRA modules, the gap between learned and uniform composition increases as well. This observation suggests that the learned composition allows the model to find a specific interpolation of LoRA modules that best suits the downstream task. Further, Figure \ref{fig:long_seq} shows that as we increase the number of splits, the performance of learned and uniform compositions decreases. This can be explained by the overfitting of upstream tasks to few labels and therefore these upstream tasks cannot generalize to the new downstream task.


\setlength{\tabcolsep}{2pt}
\begin{table}[h!]
  \small
  \centering
  \resizebox{\columnwidth}{!}{
  \begin{tabular}{llccccc}
    \toprule
        \multirow{2}{*}{\begin{tabular}{@{}c@{}}Downstream \\ Classes\end{tabular}} & \multirow{2}{*}{\begin{tabular}{@{}c@{}}Composition \\ Method\end{tabular}} & \multicolumn{4}{c}{Weights of Upstream Tasks} & \multirow{2}{*}{Acc. (\%)}\\
        \cmidrule(lr){3-6}
         & & $v_{1 - 25}$ & $v_{26 - 50}$ & $v_{51 - 75}$ & $v_{76 - 100}$\\
        
         \midrule

         \multirow{2}{*}{75 - 100} & uniform & 0.25 & 0.25 & 0.25 & 0.25 & 95.7 \\
         & learned & 0.08 & 0.09 & 0.15 & 0.68 & \textbf{98.9} \\
        \midrule
         \multirow{2}{*}{25 - 50} & uniform & 0.25 & 0.25 & 0.25 & 0.25 & 94.1 \\
         & learned & 0.07 & 0.67 & 0.14 & 0.10 & \textbf{96.3} \\
        \midrule
         \multirow{2}{*}{15 - 40} & uniform & 0.25 & 0.25 & 0.25 & 0.25 & 93.8 \\
         & learned & 0.24 & 0.46 & 0.19 & 0.11 & \textbf{95.9} \\
         
    \bottomrule
  \end{tabular}}
  \caption{Ablation on entangled upstream and downstream data under label shift setting on Split-CIFAR100. \vspace{-1em}}
  \label{tab:entangling}
\end{table}

\subsection{Entangling downstream and upstream tasks}
In our previous experiments, we assumed that the upstream tasks were completely independent of the downstream task. However, in \cref{tab:entangling}, we explore a scenario where one or more upstream tasks have partial data relevant to the downstream task. More specifically, we utilize the Split-CIFAR100 dataset with 4 disjoint upstream tasks of 25 labels, where upstream tasks have access to labels 1-25, 26-50, 51-75, and 76-100 respectively. We evaluate three ranges of labels for the downstream task: (a) 76-100, (b) 26-50, and (d) 15-40.
Looking at the results in \cref{tab:entangling}, we notice a strong correlation between the similarity of the downstream and upstream tasks and the merging weights generated by the learned composition method. This finding implies that the learned composition method can effectively select upstream modules that provide the most benefit to the downstream task.

\setlength{\tabcolsep}{4pt}
\begin{table}[h!]
  \small
  \centering
  \begin{tabular}{l cccc}
    \toprule
    \multirow{2}{*}{Method} &   \multicolumn{4}{c}{Food101}\\
    \cmidrule{2-5}
    & r = 4 & r = 8 & r = 16 & r = 32\\
    \midrule

    LoRA & 84.21\tiny{ $\pm$ 0.17} & 84.24\tiny{ $\pm$ 0.12} & 84.19\tiny{ $\pm$ 0.10} & 84.03\tiny{ $\pm$ 0.10} \\
    uniform & 79.64\tiny{ $\pm$ 0.03} & 79.73\tiny{ $\pm$ 0.04} & 79.81\tiny{ $\pm$ 0.04} & 79.70\tiny{ $\pm$ 0.04} \\
    learned & 81.97\tiny{ $\pm$ 0.08} & 82.22\tiny{ $\pm$ 0.10} & 82.07\tiny{ $\pm$0.09} & 81.88\tiny{ $\pm$ 0.10} \\

   \bottomrule
  \end{tabular}
  \vspace{-0.0 em}
  \caption{\textbf{Ablation on rank}. We validate the downstream accuracy on Food101 dataset under task shift setting. }
  \vspace{-.5 em}
  \label{tab:rank}
\end{table}

\subsection{Effect of Rank ($r$)}
So far, in the previous experiments we set the rank $r=16$. In this section, we evaluate the effect of $r$ on the combination methods using the task shift settings. More specifically, we compare module combination methods for a rank of 4, 8, 16, and 32. \cref{tab:rank} shows that surprisingly the performance of the LoRA baseline as well as the composition methods are very competitive with small rank. This observation is consistent with the ablation study in LoRA rank in the original paper \cite{hu2021lora}.


\subsection{Visualizing Learned Composition}\label{sec:cka}

\begin{figure}
    \begin{minipage}{0.254\textwidth}
        \includegraphics[width=\textwidth]{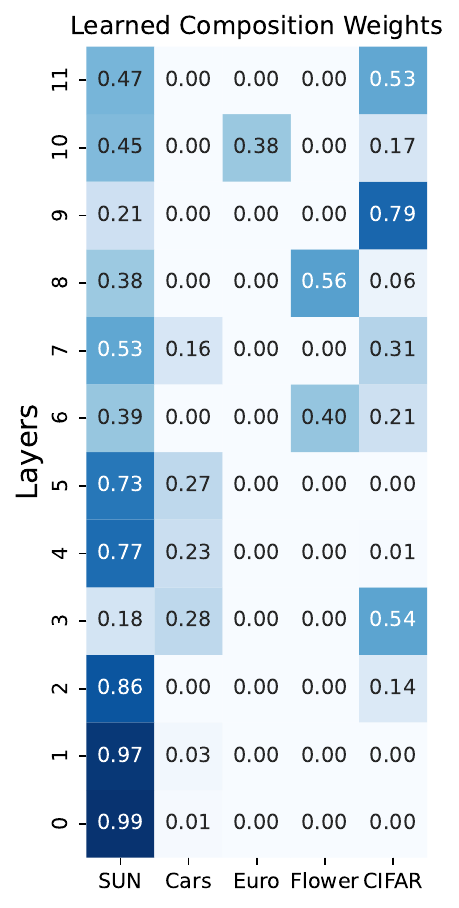}
    \end{minipage}\hfill
    \begin{minipage}{0.225\textwidth}
        \includegraphics[width=\textwidth]{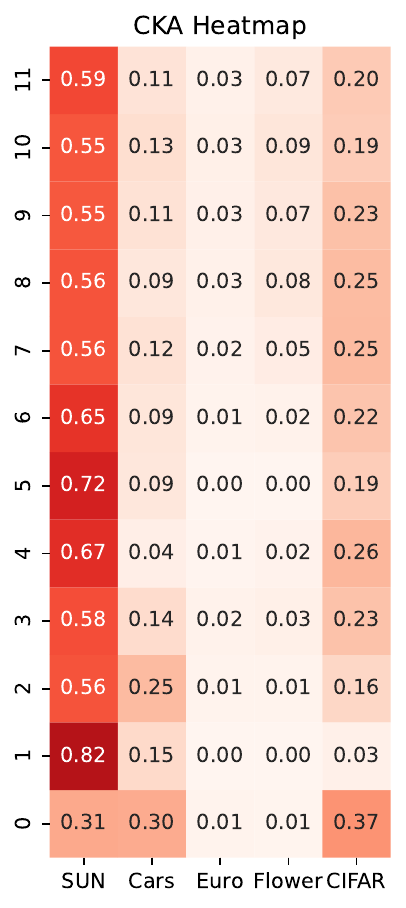}
    \end{minipage}\hfill
    \vspace{-0.5em}
    \caption{Visualization of the learned composition weights $\mathbf{v}$~(left) and the CKA similarity map~(right) for the \emph{key} weight matrix of ViT-base. The x-axis represents the upstream LoRA module and the y-axis represents the layer number. We can observe a high correlation between the upstream module picked by learned composition and the CKA similarity of upstream and downstream.}

    \label{fig:cka}
\end{figure}

This section studies the internal representation structure of the learned LoRA composition and see if the learned composition selects the \textit{relevant} upstream module. Our exploration involves visualizing the learned vector $\mathbf{v}$ at each layer of the network. At the same time, to measure the relevancy between the downstream task and each upstream task,  we use Canonical Correlation Analysis (CKA)~\cite{kornblith2019better} to quantify the feature similarity between pre-trained upstream LoRA modules and the ground-truth LoRA module of the downstream task. CKA is a widely adopted metric for assessing feature representation similarity across layers and models trained on diverse tasks~\cite{ramasesh2020anatomy, davari2022probing}. Given a dataset with $m$ samples and their representations denoted as $X$ and $Y$, the linear CKA between $X$ and $Y$ is expressed as:
\begin{equation*}
    \text{CKA}(XX^\top, YY^\top) = \frac{\|Y^{\top} X\|^2_F}{\|X^{\top} X\|_F \|Y^{\top} Y\|_F} 
\end{equation*}
Consequently, in our analysis, a high CKA value indicates a significant similarity in representations between the upstream and downstream tasks at a specific layer of the network. In Figure \ref{fig:cka}, we present visualizations of the learned composition vectors $\mathbf{v}$ and the CKA heatmap for the ``key" weight matrices of attention modules across all layers of the ViT-base model. For this analysis, Food101 serves as the downstream task, while Stanford Cars, SUN397, Eurosat, CIFAR100, and Flowers are selected as the upstream tasks. The CKA values are normalized across the upstreams~(x-axis). Notably, our observations reveal a preference for utilizing SUN397 and CIFAR100 upstream modules in the learned composition. Intriguingly, the CKA heatmap aligns with this pattern, indicating a higher similarity between the feature embeddings of the upstream tasks (SUN397 and Stanford Cars) and the Food101 downstream task. This alignment suggests that the learned composition effectively combines valuable information from the upstream tasks.

\section{Conclusion}
Our investigation into the composability of LoRA modules demonstrates their efficacy in enhancing transferability for downstream tasks. Both uniform and learned composition approaches prove advantageous, particularly in few-shot settings, surpassing traditional fine-tuning methods and even outperforming training a LoRA from scratch by up to 10.23\%. This research underscores the potential of uniform composition for improving transfer accuracy in low-shot settings without introducing additional learnable parameters.



\bibliography{references}

\begin{thebibliography}{43}
\providecommand{\natexlab}[1]{#1}
\providecommand{\url}[1]{\texttt{#1}}
\expandafter\ifx\csname urlstyle\endcsname\relax
  \providecommand{\doi}[1]{doi: #1}\else
  \providecommand{\doi}{doi: \begingroup \urlstyle{rm}\Url}\fi

\bibitem[Bapna et~al.(2019)Bapna, Arivazhagan, and Firat]{bapna2019simple}
Bapna, A., Arivazhagan, N., and Firat, O.
\newblock Simple, scalable adaptation for neural machine translation.
\newblock \emph{arXiv preprint arXiv:1909.08478}, 2019.

\bibitem[Bhakthavatsalam et~al.(2021)Bhakthavatsalam, Khashabi, Khot, Mishra, Richardson, Sabharwal, Schoenick, Tafjord, and Clark]{bhakthavatsalam2021think}
Bhakthavatsalam, S., Khashabi, D., Khot, T., Mishra, B.~D., Richardson, K., Sabharwal, A., Schoenick, C., Tafjord, O., and Clark, P.
\newblock Think you have solved direct-answer question answering? try arc-da, the direct-answer ai2 reasoning challenge.
\newblock \emph{arXiv preprint arXiv:2102.03315}, 2021.

\bibitem[Bommasani et~al.(2021)Bommasani, Hudson, Adeli, Altman, Arora, von Arx, Bernstein, Bohg, Bosselut, Brunskill, et~al.]{bommasani2021opportunities}
Bommasani, R., Hudson, D.~A., Adeli, E., Altman, R., Arora, S., von Arx, S., Bernstein, M.~S., Bohg, J., Bosselut, A., Brunskill, E., et~al.
\newblock On the opportunities and risks of foundation models.
\newblock \emph{arXiv preprint arXiv:2108.07258}, 2021.

\bibitem[Bossard et~al.(2014)Bossard, Guillaumin, and Van~Gool]{bossard2014food}
Bossard, L., Guillaumin, M., and Van~Gool, L.
\newblock Food-101--mining discriminative components with random forests.
\newblock In \emph{Computer Vision--ECCV 2014: 13th European Conference, Zurich, Switzerland, September 6-12, 2014, Proceedings, Part VI 13}, pp.\  446--461. Springer, 2014.

\bibitem[Caccia et~al.(2023)Caccia, Ponti, Su, Pereira, Le~Roux, and Sordoni]{caccia2023multi}
Caccia, L., Ponti, E., Su, Z., Pereira, M., Le~Roux, N., and Sordoni, A.
\newblock Multi-head adapter routing for cross-task generalization.
\newblock In \emph{Thirty-seventh Conference on Neural Information Processing Systems}, 2023.

\bibitem[Choshen et~al.(2022)Choshen, Venezian, Slonim, and Katz]{choshen2022fusing}
Choshen, L., Venezian, E., Slonim, N., and Katz, Y.
\newblock Fusing finetuned models for better pretraining.
\newblock \emph{arXiv preprint arXiv:2204.03044}, 2022.

\bibitem[Chung et~al.(2022)Chung, Hou, Longpre, Zoph, Tay, Fedus, Li, Wang, Dehghani, Brahma, et~al.]{chung2022scaling}
Chung, H.~W., Hou, L., Longpre, S., Zoph, B., Tay, Y., Fedus, W., Li, Y., Wang, X., Dehghani, M., Brahma, S., et~al.
\newblock Scaling instruction-finetuned language models.
\newblock \emph{arXiv preprint arXiv:2210.11416}, 2022.

\bibitem[Davari \& Belilovsky(2023)Davari and Belilovsky]{davari2023model}
Davari, M. and Belilovsky, E.
\newblock Model breadcrumbs: Scaling multi-task model merging with sparse masks.
\newblock \emph{arXiv preprint arXiv:2312.06795}, 2023.

\bibitem[Davari et~al.(2022)Davari, Asadi, Mudur, Aljundi, and Belilovsky]{davari2022probing}
Davari, M., Asadi, N., Mudur, S., Aljundi, R., and Belilovsky, E.
\newblock Probing representation forgetting in supervised and unsupervised continual learning.
\newblock In \emph{Proceedings of the IEEE/CVF Conference on Computer Vision and Pattern Recognition}, pp.\  16712--16721, 2022.

\bibitem[Dettmers et~al.(2023)Dettmers, Pagnoni, Holtzman, and Zettlemoyer]{dettmers2023qlora}
Dettmers, T., Pagnoni, A., Holtzman, A., and Zettlemoyer, L.
\newblock Qlora: Efficient finetuning of quantized llms.
\newblock \emph{arXiv preprint arXiv:2305.14314}, 2023.

\bibitem[Dosovitskiy et~al.(2020)Dosovitskiy, Beyer, Kolesnikov, Weissenborn, Zhai, Unterthiner, Dehghani, Minderer, Heigold, Gelly, et~al.]{dosovitskiy2020image}
Dosovitskiy, A., Beyer, L., Kolesnikov, A., Weissenborn, D., Zhai, X., Unterthiner, T., Dehghani, M., Minderer, M., Heigold, G., Gelly, S., et~al.
\newblock An image is worth 16x16 words: Transformers for image recognition at scale.
\newblock \emph{arXiv preprint arXiv:2010.11929}, 2020.

\bibitem[Gandikota et~al.(2023)Gandikota, Materzynska, Zhou, Torralba, and Bau]{gandikota2023concept}
Gandikota, R., Materzynska, J., Zhou, T., Torralba, A., and Bau, D.
\newblock Concept sliders: Lora adaptors for precise control in diffusion models.
\newblock \emph{arXiv preprint arXiv:2311.12092}, 2023.

\bibitem[Guo et~al.(2020)Guo, Rush, and Kim]{guo2020parameter}
Guo, D., Rush, A.~M., and Kim, Y.
\newblock Parameter-efficient transfer learning with diff pruning.
\newblock \emph{arXiv preprint arXiv:2012.07463}, 2020.

\bibitem[Helber et~al.(2019)Helber, Bischke, Dengel, and Borth]{helber2019eurosat}
Helber, P., Bischke, B., Dengel, A., and Borth, D.
\newblock Eurosat: A novel dataset and deep learning benchmark for land use and land cover classification.
\newblock \emph{IEEE Journal of Selected Topics in Applied Earth Observations and Remote Sensing}, 12\penalty0 (7):\penalty0 2217--2226, 2019.

\bibitem[Hu et~al.(2021)Hu, Shen, Wallis, Allen-Zhu, Li, Wang, Wang, and Chen]{hu2021lora}
Hu, E.~J., Shen, Y., Wallis, P., Allen-Zhu, Z., Li, Y., Wang, S., Wang, L., and Chen, W.
\newblock Lora: Low-rank adaptation of large language models.
\newblock \emph{arXiv preprint arXiv:2106.09685}, 2021.

\bibitem[Huang et~al.(2023)Huang, Liu, Lin, Pang, Du, and Lin]{huang2023lorahub}
Huang, C., Liu, Q., Lin, B.~Y., Pang, T., Du, C., and Lin, M.
\newblock Lorahub: Efficient cross-task generalization via dynamic lora composition.
\newblock \emph{arXiv preprint arXiv:2307.13269}, 2023.

\bibitem[Ilharco et~al.(2022)Ilharco, Ribeiro, Wortsman, Gururangan, Schmidt, Hajishirzi, and Farhadi]{ilharco2022editing}
Ilharco, G., Ribeiro, M.~T., Wortsman, M., Gururangan, S., Schmidt, L., Hajishirzi, H., and Farhadi, A.
\newblock Editing models with task arithmetic.
\newblock \emph{arXiv preprint arXiv:2212.04089}, 2022.

\bibitem[Kornblith et~al.(2019)Kornblith, Shlens, and Le]{kornblith2019better}
Kornblith, S., Shlens, J., and Le, Q.~V.
\newblock Do better imagenet models transfer better?
\newblock In \emph{Proceedings of the IEEE/CVF conference on computer vision and pattern recognition}, pp.\  2661--2671, 2019.

\bibitem[Krause et~al.(2013)Krause, Stark, Deng, and Fei-Fei]{krause20133d}
Krause, J., Stark, M., Deng, J., and Fei-Fei, L.
\newblock 3d object representations for fine-grained categorization.
\newblock In \emph{Proceedings of the IEEE international conference on computer vision workshops}, pp.\  554--561, 2013.

\bibitem[Krizhevsky et~al.(2009)Krizhevsky, Hinton, et~al.]{krizhevsky2009learning}
Krizhevsky, A., Hinton, G., et~al.
\newblock Learning multiple layers of features from tiny images.
\newblock 2009.

\bibitem[Lester et~al.(2021)Lester, Al-Rfou, and Constant]{lester2021power}
Lester, B., Al-Rfou, R., and Constant, N.
\newblock The power of scale for parameter-efficient prompt tuning.
\newblock \emph{arXiv preprint arXiv:2104.08691}, 2021.

\bibitem[Li et~al.(2017)Li, Yang, Song, and Hospedales]{li2017deeper}
Li, D., Yang, Y., Song, Y.-Z., and Hospedales, T.~M.
\newblock Deeper, broader and artier domain generalization.
\newblock In \emph{Proceedings of the IEEE international conference on computer vision}, pp.\  5542--5550, 2017.

\bibitem[Li \& Liang(2021)Li and Liang]{li2021prefix}
Li, X.~L. and Liang, P.
\newblock Prefix-tuning: Optimizing continuous prompts for generation.
\newblock \emph{arXiv preprint arXiv:2101.00190}, 2021.

\bibitem[Lin et~al.(2020)Lin, Madotto, and Fung]{lin2020exploring}
Lin, Z., Madotto, A., and Fung, P.
\newblock Exploring versatile generative language model via parameter-efficient transfer learning.
\newblock In \emph{Findings of the Association for Computational Linguistics: EMNLP 2020}, pp.\  441--459, 2020.

\bibitem[Liu et~al.(2022)Liu, Tam, Muqeeth, Mohta, Huang, Bansal, and Raffel]{liu2022few}
Liu, H., Tam, D., Muqeeth, M., Mohta, J., Huang, T., Bansal, M., and Raffel, C.~A.
\newblock Few-shot parameter-efficient fine-tuning is better and cheaper than in-context learning.
\newblock \emph{Advances in Neural Information Processing Systems}, 35:\penalty0 1950--1965, 2022.

\bibitem[Matena \& Raffel()Matena and Raffel]{matena2111merging}
Matena, M. and Raffel, C.
\newblock Merging models with fisher-weighted averaging, 2021.
\newblock \emph{arXiv preprint arXiv:2111.09832}.

\bibitem[Nilsback \& Zisserman(2008)Nilsback and Zisserman]{nilsback2008automated}
Nilsback, M.-E. and Zisserman, A.
\newblock Automated flower classification over a large number of classes.
\newblock In \emph{2008 Sixth Indian conference on computer vision, graphics \& image processing}, pp.\  722--729. IEEE, 2008.

\bibitem[Ponti et~al.(2023)Ponti, Sordoni, Bengio, and Reddy]{ponti2023combining}
Ponti, E.~M., Sordoni, A., Bengio, Y., and Reddy, S.
\newblock Combining parameter-efficient modules for task-level generalisation.
\newblock In \emph{Proceedings of the 17th Conference of the European Chapter of the Association for Computational Linguistics}, pp.\  687--702, 2023.

\bibitem[Radford et~al.(2021)Radford, Kim, Hallacy, Ramesh, Goh, Agarwal, Sastry, Askell, Mishkin, Clark, et~al.]{radford2021learning}
Radford, A., Kim, J.~W., Hallacy, C., Ramesh, A., Goh, G., Agarwal, S., Sastry, G., Askell, A., Mishkin, P., Clark, J., et~al.
\newblock Learning transferable visual models from natural language supervision.
\newblock In \emph{International conference on machine learning}, pp.\  8748--8763. PMLR, 2021.

\bibitem[Raffel et~al.(2020)Raffel, Shazeer, Roberts, Lee, Narang, Matena, Zhou, Li, and Liu]{raffel2020exploring}
Raffel, C., Shazeer, N., Roberts, A., Lee, K., Narang, S., Matena, M., Zhou, Y., Li, W., and Liu, P.~J.
\newblock Exploring the limits of transfer learning with a unified text-to-text transformer.
\newblock \emph{The Journal of Machine Learning Research}, 21\penalty0 (1):\penalty0 5485--5551, 2020.

\bibitem[Ramasesh et~al.(2020)Ramasesh, Dyer, and Raghu]{ramasesh2020anatomy}
Ramasesh, V.~V., Dyer, E., and Raghu, M.
\newblock Anatomy of catastrophic forgetting: Hidden representations and task semantics.
\newblock \emph{arXiv preprint arXiv:2007.07400}, 2020.

\bibitem[Ram{\'e} et~al.(2023)Ram{\'e}, Ahuja, Zhang, Cord, Bottou, and Lopez-Paz]{rame2023model}
Ram{\'e}, A., Ahuja, K., Zhang, J., Cord, M., Bottou, L., and Lopez-Paz, D.
\newblock Model ratatouille: Recycling diverse models for out-of-distribution generalization.
\newblock In \emph{International Conference on Machine Learning}, pp.\  28656--28679. PMLR, 2023.

\bibitem[Rebuffi et~al.(2017)Rebuffi, Bilen, and Vedaldi]{rebuffi2017learning}
Rebuffi, S.-A., Bilen, H., and Vedaldi, A.
\newblock Learning multiple visual domains with residual adapters.
\newblock \emph{Advances in neural information processing systems}, 30, 2017.

\bibitem[Ridnik et~al.(2021)Ridnik, Ben-Baruch, Noy, and Zelnik-Manor]{ridnik2021imagenet}
Ridnik, T., Ben-Baruch, E., Noy, A., and Zelnik-Manor, L.
\newblock Imagenet-21k pretraining for the masses.
\newblock \emph{arXiv preprint arXiv:2104.10972}, 2021.

\bibitem[Rogers et~al.(2020)Rogers, Kovaleva, Downey, and Rumshisky]{rogers2020getting}
Rogers, A., Kovaleva, O., Downey, M., and Rumshisky, A.
\newblock Getting closer to ai complete question answering: A set of prerequisite real tasks.
\newblock In \emph{Proceedings of the AAAI conference on artificial intelligence}, volume~34, pp.\  8722--8731, 2020.

\bibitem[Shah et~al.(2023)Shah, Ruiz, Cole, Lu, Lazebnik, Li, and Jampani]{shah2023ziplora}
Shah, V., Ruiz, N., Cole, F., Lu, E., Lazebnik, S., Li, Y., and Jampani, V.
\newblock Ziplora: Any subject in any style by effectively merging loras.
\newblock \emph{arXiv preprint arXiv:2311.13600}, 2023.

\bibitem[Sung et~al.(2021)Sung, Nair, and Raffel]{sung2021training}
Sung, Y.-L., Nair, V., and Raffel, C.~A.
\newblock Training neural networks with fixed sparse masks.
\newblock \emph{Advances in Neural Information Processing Systems}, 34:\penalty0 24193--24205, 2021.

\bibitem[Touvron et~al.(2023)Touvron, Martin, Stone, Albert, Almahairi, Babaei, Bashlykov, Batra, Bhargava, Bhosale, et~al.]{touvron2023llama}
Touvron, H., Martin, L., Stone, K., Albert, P., Almahairi, A., Babaei, Y., Bashlykov, N., Batra, S., Bhargava, P., Bhosale, S., et~al.
\newblock Llama 2: Open foundation and fine-tuned chat models.
\newblock \emph{arXiv preprint arXiv:2307.09288}, 2023.

\bibitem[Welbl et~al.(2017)Welbl, Liu, and Gardner]{welbl2017crowdsourcing}
Welbl, J., Liu, N.~F., and Gardner, M.
\newblock Crowdsourcing multiple choice science questions.
\newblock \emph{arXiv preprint arXiv:1707.06209}, 2017.

\bibitem[Wortsman et~al.(2022)Wortsman, Ilharco, Gadre, Roelofs, Gontijo-Lopes, Morcos, Namkoong, Farhadi, Carmon, Kornblith, et~al.]{wortsman2022model}
Wortsman, M., Ilharco, G., Gadre, S.~Y., Roelofs, R., Gontijo-Lopes, R., Morcos, A.~S., Namkoong, H., Farhadi, A., Carmon, Y., Kornblith, S., et~al.
\newblock Model soups: averaging weights of multiple fine-tuned models improves accuracy without increasing inference time.
\newblock In \emph{International Conference on Machine Learning}, pp.\  23965--23998. PMLR, 2022.

\bibitem[Xiao et~al.(2010)Xiao, Hays, Ehinger, Oliva, and Torralba]{xiao2010sun}
Xiao, J., Hays, J., Ehinger, K.~A., Oliva, A., and Torralba, A.
\newblock Sun database: Large-scale scene recognition from abbey to zoo.
\newblock In \emph{2010 IEEE computer society conference on computer vision and pattern recognition}, pp.\  3485--3492. IEEE, 2010.

\bibitem[Xu et~al.(2023)Xu, Guo, Duan, and McAuley]{xu2023baize}
Xu, C., Guo, D., Duan, N., and McAuley, J.
\newblock Baize: An open-source chat model with parameter-efficient tuning on self-chat data.
\newblock \emph{arXiv preprint arXiv:2304.01196}, 2023.

\bibitem[Ye et~al.(2021)Ye, Lin, and Ren]{ye2021crossfit}
Ye, Q., Lin, B.~Y., and Ren, X.
\newblock Crossfit: A few-shot learning challenge for cross-task generalization in nlp.
\newblock \emph{arXiv preprint arXiv:2104.08835}, 2021.

\end{thebibliography}
\bibliographystyle{icml2024}

\newpage
\appendix
\onecolumn
\section{Results}
In this section, we provide our complete experimental results for task and domain shift settings in table format.

\setlength{\tabcolsep}{6pt} 
\renewcommand{\arraystretch}{1.2} 

\subsection{Task Shift Results}

\begin{table*}[h!]
  \small
  \centering
  \begin{tabular}{l cccccccc}
    \toprule
    \multirow{2}{*}{Method} &   \multicolumn{7}{c}{Food101} &\multirow{2}{*}{$|\Theta|$}\\
    \cline{2-8}
    & K=1 & K=2 & K=5 & K=10 & K=20 & K=50  & K=all & \\
    \midrule

    Classifier Tuning  &30.94\tiny{$\pm$ 1.62}  &40.21\tiny{$\pm$ 1.59}  &51.77\tiny{$\pm$ 0.68}  &58.43\tiny{$\pm$ 0.11}  &63.17\tiny{$\pm$ 0.16}  &67.52\tiny{$\pm$ 0.25}  &75.89\tiny{$\pm$ 0.03}  & 0  \\
    Full Fine-tuning  &32.92\tiny{$\pm$ 1.67}  &43.31\tiny{$\pm$ 1.47}  &59.29\tiny{$\pm$ 0.57}  &\underline{67.25}\tiny{$\pm$ 0.21}  &\textbf{72.40}\tiny{$\pm$ 0.32}  &\textbf{76.36}\tiny{$\pm$ 0.20}  &\textbf{84.23}\tiny{$\pm$ 0.15}  & 86M  \\
    LoRA  &31.02\tiny{$\pm$ 1.72}  &41.30\tiny{$\pm$ 1.80}  &55.71\tiny{$\pm$ 0.54}  &64.40\tiny{$\pm$ 0.28}  &70.19\tiny{$\pm$ 0.14}  &75.27\tiny{$\pm$ 0.19}  &\underline{83.76}\tiny{$\pm$ 0.64}  &  0.88M  \\
    \midrule
    Uniform Composition  &\underline{33.71}\tiny{$\pm$ 0.26}  &\underline{47.50}\tiny{$\pm$ 0.37}  &\underline{60.06}\tiny{$\pm$ 0.50}  &65.50\tiny{$\pm$ 0.12}  &69.33\tiny{$\pm$ 0.06}  &72.78\tiny{$\pm$ 0.23}  &79.93\tiny{$\pm$ 0.04}  & 0  \\
    Learned Composition  &\textbf{36.09}\tiny{$\pm$ 0.24}  &\textbf{49.88}\tiny{$\pm$ 0.50}  &\textbf{63.03}\tiny{$\pm$ 0.61}  &\textbf{68.37}\tiny{$\pm$ 0.49}  &\underline{72.03}\tiny{$\pm$ 0.79}  &\underline{76.26}\tiny{$\pm$ 0.57}  &82.31\tiny{$\pm$ 0.49}  &  108  \\

   \bottomrule
  \end{tabular}

  \label{tab:sup_food}
\end{table*}

\begin{table*}[h!]
  \small
  \centering
  \begin{tabular}{l cccccccc}
    \toprule
    \multirow{2}{*}{Method} &   \multicolumn{7}{c}{Eurosat} & \multirow{2}{*}{$|\Theta|$}\\
    \cline{2-8}
    & K=1 & K=2 & K=5 & K=10 & K=20 & K=50  & K=all \\
    \midrule

    Classifier Tuning  &43.88\tiny{$\pm$ 1.62}  &58.27\tiny{$\pm$ 0.83}  &70.98\tiny{$\pm$ 1.48}  &75.50\tiny{$\pm$ 0.23}  &78.32\tiny{$\pm$ 1.00}  &85.78\tiny{$\pm$ 0.28}  &95.72\tiny{$\pm$ 0.05}  & 0  \\
    Full Fine-tuning  &\underline{53.14}\tiny{$\pm$ 3.39}  &\underline{68.58}\tiny{$\pm$ 0.78}  &83.41\tiny{$\pm$ 0.13}  &\textbf{87.65}\tiny{$\pm$ 0.35}  &\textbf{92.56}\tiny{$\pm$ 0.10}  &\textbf{95.37}\tiny{$\pm$ 0.34}  &\textbf{98.84}\tiny{$\pm$ 0.03}  & 86M  \\
    LoRA  &44.37\tiny{$\pm$ 2.03}  &58.21\tiny{$\pm$ 0.14}  &72.29\tiny{$\pm$ 1.76}  &76.96\tiny{$\pm$ 0.44}  &80.05\tiny{$\pm$ 0.64}  &88.54\tiny{$\pm$ 0.40}  &\underline{98.37}\tiny{$\pm$ 0.02}  & 0.88M  \\
    \midrule
    Uniform Composition  &52.20\tiny{$\pm$ 4.02}  &66.86\tiny{$\pm$ 3.30}  &80.23\tiny{$\pm$ 0.42}  &82.24\tiny{$\pm$ 0.41}  &85.32\tiny{$\pm$ 0.59}  &90.76\tiny{$\pm$ 0.20}  &96.68\tiny{$\pm$ 0.30}  & 0  \\
    Learned Composition  &\textbf{55.68}\tiny{$\pm$ 4.24}  &\textbf{70.53}\tiny{$\pm$ 3.47}  &\textbf{85.16}\tiny{$\pm$ 0.66}  &\underline{87.50}\tiny{$\pm$ 0.71}  &\underline{91.13}\tiny{$\pm$ 0.27}  &\underline{95.21}\tiny{$\pm$ 0.12}  &98.05\tiny{$\pm$ 0.46}  &  108  \\

   \bottomrule
  \end{tabular}
  
  \label{tab:sup_eurosat}
\end{table*}

\begin{table*}[h!]
  \small
  \centering
  \begin{tabular}{l cccccccc}
    \toprule
    \multirow{2}{*}{Method} &   \multicolumn{7}{c}{Flowers} & \multirow{2}{*}{$|\Theta|$}\\
    \cline{2-8}
    & K=1 & K=2 & K=5 & K=10 & K=20 & K=50  & K=all \\
    \midrule

    Classifier Tuning  &87.49\tiny{$\pm$ 0.82}  &90.09\tiny{$\pm$ 0.67}  &93.85\tiny{$\pm$ 0.73}  &95.45\tiny{$\pm$ 0.22}  &95.45\tiny{$\pm$ 0.22}  &95.45\tiny{$\pm$ 0.22}  &95.62\tiny{$\pm$ 0.07}  & 0  \\
    Full Fine-tuning  &89.23\tiny{$\pm$ 0.41}  &93.26\tiny{$\pm$ 0.85}  &\textbf{97.22}\tiny{$\pm$ 0.38}  &\textbf{98.02}\tiny{$\pm$ 0.19}  &\textbf{98.02}\tiny{$\pm$ 0.19}  &\textbf{98.02}\tiny{$\pm$ 0.19}  &\textbf{98.42}\tiny{$\pm$ 0.02}  & 86M  \\
    LoRA  &87.83\tiny{$\pm$ 1.09}  &90.63\tiny{$\pm$ 0.80}  &95.24\tiny{$\pm$ 0.49}  &96.97\tiny{$\pm$ 0.21}  &96.97\tiny{$\pm$ 0.21}  &97.07\tiny{$\pm$ 0.26}  &97.55\tiny{$\pm$ 0.04}  & 0.88M  \\
    \midrule
    Uniform Composition  &\underline{91.29}\tiny{$\pm$ 0.21}  &\underline{94.35}\tiny{$\pm$ 0.59}  &96.46\tiny{$\pm$ 0.53}  &97.40\tiny{$\pm$ 0.19}  &97.37\tiny{$\pm$ 0.17}  &97.39\tiny{$\pm$ 0.17}  &97.63\tiny{$\pm$ 0.07}  & 0  \\
    Learned Composition  &\textbf{91.95}\tiny{$\pm$ 0.46}  &\textbf{94.80}\tiny{$\pm$ 0.58}  &\underline{96.86}\tiny{$\pm$ 0.50}  &\underline{97.75}\tiny{$\pm$ 0.11}  &\underline{97.75}\tiny{$\pm$ 0.11}  &\underline{97.81}\tiny{$\pm$ 0.09}  &\underline{98.08}\tiny{$\pm$ 0.08}  & 108  \\

   \bottomrule
  \end{tabular}

  \label{tab:sup_flowers}
\end{table*}

\begin{table*}[h!]
  \small
  \centering
  \begin{tabular}{l cccccccc}
    \toprule
    \multirow{2}{*}{Method} &   \multicolumn{7}{c}{CIFAR100} & \multirow{2}{*}{$|\Theta|$}\\
    \cline{2-8}
    & K=1 & K=2 & K=5 & K=10 & K=20 & K=50  & K=all \\
    \midrule

    Classifier Tuning  &41.37\tiny{$\pm$ 1.00}  &50.71\tiny{$\pm$ 2.17}  &60.69\tiny{$\pm$ 0.41}  &66.54\tiny{$\pm$ 0.16}  &69.86\tiny{$\pm$ 0.19}  &72.69\tiny{$\pm$ 0.09}  &77.70\tiny{$\pm$ 0.02}  & 0  \\
    Full Fine-tuning  &46.74\tiny{$\pm$ 0.80}  &\underline{59.93}\tiny{$\pm$ 1.53}  &\textbf{74.53}\tiny{$\pm$ 0.92}  &\textbf{82.26}\tiny{$\pm$ 0.37}  &\textbf{85.26}\tiny{$\pm$ 0.25}  &\textbf{86.73}\tiny{$\pm$ 0.22}  &\textbf{89.97}\tiny{$\pm$ 0.04}  & 86M  \\
    LoRA  &41.59\tiny{$\pm$ 0.92}  &51.92\tiny{$\pm$ 2.17}  &65.37\tiny{$\pm$ 0.68}  &75.59\tiny{$\pm$ 0.18}  &\underline{80.93}\tiny{$\pm$ 0.29}  &\underline{84.36}\tiny{$\pm$ 0.18}  &\underline{89.76}\tiny{$\pm$ 0.06}  & 0.88M  \\
    \midrule
    Uniform Composition  &\underline{49.65}\tiny{$\pm$ 0.39}  &59.20\tiny{$\pm$ 1.61}  &68.81\tiny{$\pm$ 0.44}  &74.02\tiny{$\pm$ 0.28}  &76.48\tiny{$\pm$ 0.23}  &78.77\tiny{$\pm$ 0.18}  &81.95\tiny{$\pm$ 0.46}  & 0  \\
    Learned Composition &\textbf{51.60}\tiny{$\pm$ 0.65}  &\textbf{62.51}\tiny{$\pm$ 2.02}  &\underline{72.74}\tiny{$\pm$ 0.41}  &\underline{77.54}\tiny{$\pm$ 1.27}  &80.30\tiny{$\pm$ 0.61}  &81.82\tiny{$\pm$ 0.59}  &84.64\tiny{$\pm$ 0.74}   & 108 \\

   \bottomrule
  \end{tabular}
  \caption{\textbf{Task shift results.} Here K represents the number of training samples for each class and $|\Theta|$ presents the total number of trainable parameters \emph{excluding the classifier head}.}
  \label{tab:sup_cifar100}
\end{table*}

\subsection{Covariate Shift Results}\label{sup:covariate_shift}

\begin{table*}[h!]
  \small
  \centering
  \begin{tabular}{l ccccccc}
    \toprule
    \multirow{2}{*}{Method} &   \multicolumn{6}{c}{Domain shift (Sketch)} & \multirow{2}{*}{$|\Theta|$}\\
    \cline{2-7}
    & K=1 & K=5 & K=10 & K=50 & K=100 & K=all & \\
    \midrule

    Classifier Tuning  & 35.41\tiny{ $\pm$ 2.93} & 48.15\tiny{ $\pm$ 2.91} &  51.29\tiny{ $\pm$ 1.30}& 56.69 \tiny{ $\pm$ 1.22}& 65.18\tiny{ $\pm$ 1.95}& 69.55\tiny{ $\pm$ 0.58} & 0\\
    Full Fine-tuning & 42.58\tiny{ $\pm$ 1.27}  & 60.04\tiny{ $\pm$ 0.92} & \textbf{66.37\tiny{ $\pm$ 4.64}} & \textbf{83.22 \tiny{ $\pm$ 1.51}} & \textbf{87.00 \tiny{ $\pm$ 0.29}} & \textbf{95.07\tiny{ $\pm$ 0.07}} & 86M\\
    
    LoRA & 40.72\tiny{ $\pm$ 3.63} & 52.69\tiny{ $\pm$ 4.15} & 56.64\tiny{ $\pm$ 6.02}&  70.82 \tiny{ $\pm$ 6.51}& 77.96 \tiny{ $\pm$ 1.21}& 88.98\tiny{ $\pm$ 0.06} & 0.88M\\

    \midrule

    Uniform Composition & 49.09\tiny{ $\pm$ 3.60} & 58.60\tiny{ $\pm$ 4.11} & 63.14\tiny{ $\pm$ 1.30} & 69.51\tiny{ $\pm$ 1.10} & 72.95\tiny{ $\pm$ 1.33} & 80.38\tiny{ $\pm$ 0.11} & 0\\

    Learned Composition & \textbf{49.34\tiny{ $\pm$ 3.85}} & \textbf{61.11\tiny{ $\pm$ 4.01}} & 63.31\tiny{ $\pm$ 1.86} & 71.63\tiny{ $\pm$ 0.48} & 74.78\tiny{ $\pm$ 0.96} & 81.06\tiny{ $\pm$ 0.22} & 108\\

   \bottomrule
  \end{tabular}
  \label{tab:sup_sketch}
\end{table*}

\begin{table*}[h!]
  \small
  \centering
  \begin{tabular}{l ccccccc}
    \toprule
    \multirow{2}{*}{Method} &   \multicolumn{6}{c}{Domain shift (Art painting)} & \multirow{2}{*}{$|\Theta|$}\\
    \cline{2-7}
    & K=1 & K=5 & K=10 & K=50 & K=100 & K=all & \\
    \midrule

    Classifier Tuning   &46.37\tiny{ $\pm$ 12.00} &69.35\tiny{ $\pm$ 2.50}  &73.02\tiny{ $\pm$ 2.19}  &82.88\tiny{ $\pm$ 0.97}  &84.68\tiny{ $\pm$ 0.37}   & 86.30\tiny{ $\pm$ 0.24} & 0\\
    Full Fine-tuning &50.28\tiny{ $\pm$ 9.88}  &79.05\tiny{ $\pm$ 2.83}  &\textbf{83.45\tiny{ $\pm$ 1.99}}  &\textbf{92.58\tiny{ $\pm$ 0.14}}  &\textbf{93.64\tiny{ $\pm$ 0.42}}  &\textbf{95.43\tiny{ $\pm$ 0.98}} & 86M\\
    LoRA &44.66\tiny{ $\pm$ 12.00}  &77.51\tiny{ $\pm$ 2.48} &83.29\tiny{ $\pm$ 1.15}  &91.93\tiny{ $\pm$ 0.64} &93.07\tiny{ $\pm$ 0.85}  & 93.88 \tiny{ $\pm$ 0.24} & 0.88M\\

    \midrule
    Uniform Composition &56.88\tiny{$\pm$ 11.02}  &78.73\tiny{$\pm$ 2.30}  &81.09\tiny{$\pm$ 1.92}  &86.47\tiny{$\pm$ 0.42}  &88.02\tiny{$\pm$ 1.21}  &90.06\tiny{$\pm$ 0.30}  &  0  \\
    Learned Composition &\textbf{57.21\tiny{$\pm$ 11.56}}  &\textbf{79.54\tiny{$\pm$ 2.31}}  &81.25\tiny{$\pm$ 1.75}  &86.55\tiny{$\pm$ 0.80}  &88.43\tiny{$\pm$ 0.99}  &90.87\tiny{$\pm$ 0.42}  & 108  \\

   \bottomrule
  \end{tabular}
  \caption{\textbf{Covariate shift results.} Here K represents the number of training samples for each class and $|\Theta|$ presents the total number of trainable parameters \emph{excluding the classifier head}.}
  \label{tab:sup_art}
\end{table*}




\subsection{Visualization of Learned Composition Weights}
\begin{figure}[h]
    \centering
    \subfigure[]{\includegraphics[width=0.50\textwidth]{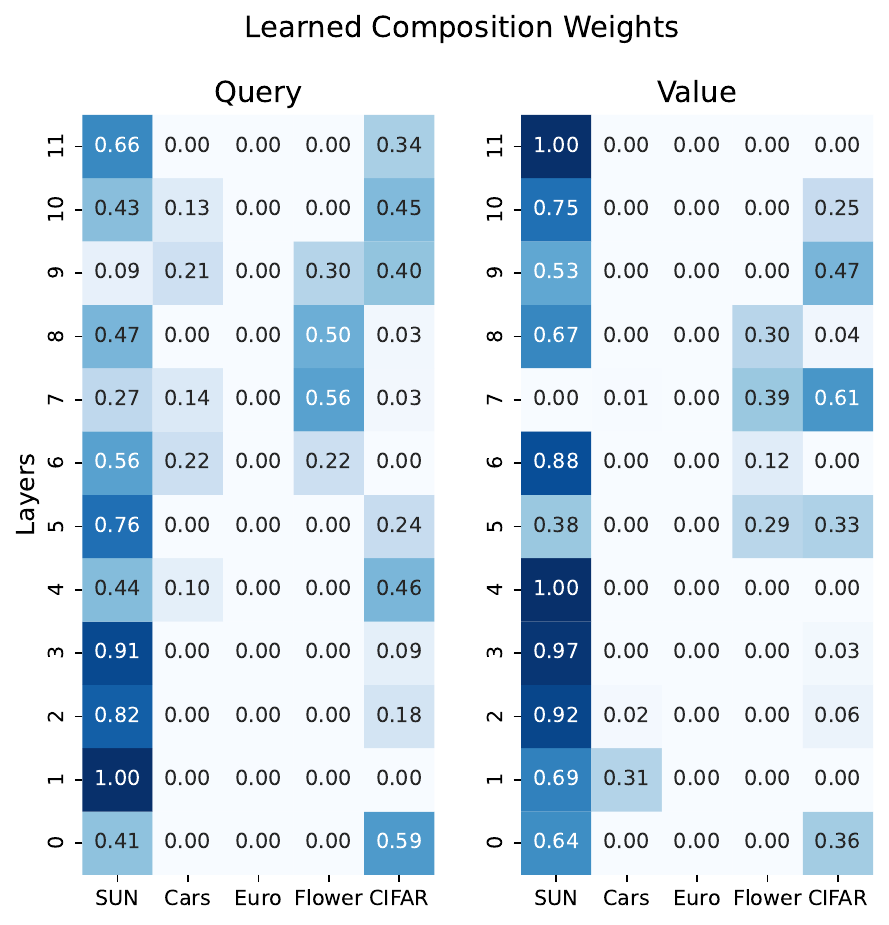}} 
    \subfigure[]{\includegraphics[width=0.48\textwidth]{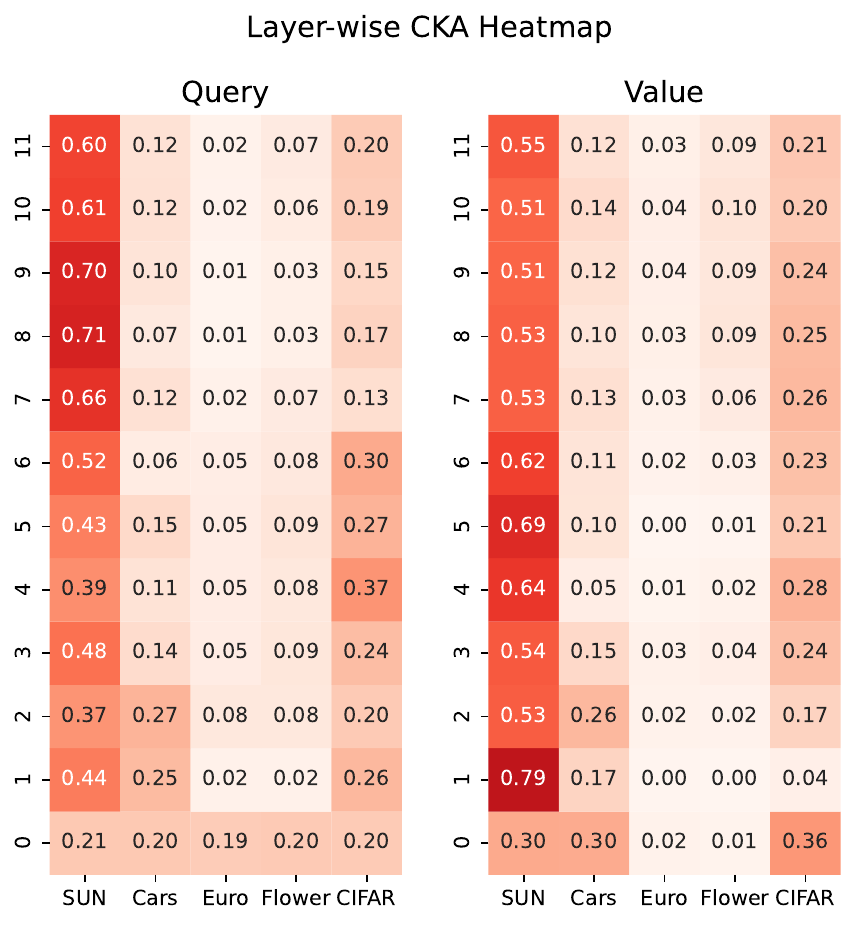}}
    \caption{ Visualization of the learned composition weights $\mathbf{v}$~(a) and the CKA similarity map~(b) for the Query and Value weight matrix of ViT-base. Here, the x-axis represents the upstream LoRA module and the y-axis represents the layer number. We can observe a high correlation between the upstream module picked by learned composition and the CKA similarity of upstream and downstream tasks.}
    \label{fig:sup_cka}
\end{figure}

In Figure \ref{fig:sup_cka}, we present visualizations of the learned composition vectors $\mathbf{v}$ and the CKA heatmap for the ``query" and ``value" weight matrices of attention modules across all layers of the ViT-base model. For this analysis, Food101 serves as the downstream task, while Stanford Cars, SUN397, Eurosat, CIFAR100, and Flowers are selected as the upstream tasks. The CKA values are normalized across the upstreams~(x-axis).

\end{document}